\documentclass{article}
\usepackage[utf8]{inputenc}

\title{Policy-Adaptable Methods For Resolving Normative Conflicts Through Argumentation and Graph Colouring}
\author{Johnny Joyce}
\date{Supervisor: Timothy Norman}

\usepackage[a4paper, total={5.5in, 8in}]{geometry}
\usepackage{multirow,array}
\usepackage{amssymb,amsmath,amsfonts,units,mathtools,amsthm}
\usepackage{hanging}
\usepackage{graphicx,subfigure,svg}
\usepackage{changepage}
\usepackage[colorlinks=false,linkcolor=blue]{hyperref}
\usepackage{microtype} % alignment improvements

\usepackage[ruled]{algorithm2e}
\usepackage{algorithmic}
\SetNlSty{bfseries}{\color{black}}{} % black line numbers

\newtheorem{proposition}{Proposition}
\newenvironment{Proof}[1][Proof]
  {\proof[#1]\leftskip=0.2cm\rightskip=0.2cm}
  {\endproof}
\newtheorem{definition}{Definition}
  
\DeclareMathOperator*{\argmax}{argmax}

\LinesNumbered

\usepackage{xcolor}

\usepackage{listings}
\definecolor{commentsColor}{rgb}{0.497495, 0.497587, 0.497464}
\definecolor{keywordsColor}{rgb}{0.000000, 0.000000, 0.635294}
\definecolor{stringColor}{rgb}{0.558215, 0.000000, 0.135316}

\begin{document}

%\maketitle

\noindent\rule{\textwidth}{1pt} % horizontal line

\begin{center}
\Large{
University of Southampton

Faculty of Engineering and Physical Sciences

Electronics and Computer Science

\vskip3em

\textit{Policy-Adaptable Methods For Resolving Normative Conflicts Through Argumentation and Graph Colouring}

by

Johnny Joyce

September 2020

\vskip3em

Supervisor: Timothy Norman

Second Examiner: Corina Cirstea

\vskip3em

A dissertation submitted in partial fulfilment of the degree

of MSc Artificial Intelligence}
\end{center}

\noindent\rule{\textwidth}{1pt} % horizontal line

\clearpage

\begin{abstract}
    In a multi-agent system, one may choose to govern the behaviour of an agent by imposing norms, which act as guidelines for how agents should act either all of the time or in given situations. However, imposing multiple norms on one or more agents may result in situations where these norms conflict over how the agent should behave. In any system with normative conflicts (such as safe reinforcement models or systems which monitor safety protocols), one must decide which norms should be followed such that the most important and most relevant norms are maintained. We introduce a new method for resolving normative conflicts through argumentation and graph colouring which is compatible with a variety of normative conflict resolution policies. We prove that this method always creates an admissible set of arguments under argumentation semantics, meaning that it produces coherent outputs. We also introduce more robust variants of this method, each building upon their predecessor to create a superior output, and we include further mathematical proof of their coherence. Our most advanced variant uses the existing concept of curtailment, where one norm may supersede another without fully eliminating it. The methods we introduce are all compatible with various pre-existing policies for resolving normative conflicts. Empirical evaluations are also performed to compare our algorithms to each other and to others in existing literature.
\end{abstract}

\vskip3em
\newpage

\section*{Statement of Originality}

------------------------------

\begin{itemize}\itemsep0em
    \item I have read and understood the ECS Academic Integrity information and the University’s Academic Integrity Guidance for Students
    \item I am aware that failure to act in accordance with the Regulations Governing Academic Integrity may lead to the imposition of penalties which, for the most serious cases, may include termination of programme.
    \item I consent to the University copying and distributing any or all of my work in any form and using third parties (who may be based outside the EU/EEA) to verify whether my work contains plagiarised material, and for quality assurance purposes.
\end{itemize}

------------------------------

\begin{itemize}
    \item I have acknowledged all sources, and identified any content taken from elsewhere.
    \item I have not used any resources produced by anyone else.
    \item I did all the work myself, or with my allocated group, and have not helped anyone else
    \item The material in the report is genuine, and I have included all my data/code/designs. 
    \item I have not submitted any part of this work for another assessment
    \item My work did not involve human participants, their cells or data, or animals.
\end{itemize}

------------------------------

\newpage

\tableofcontents

\newpage

\section{Introduction}\label{sectionintroduction}

% Non-monotonic logic \cite{dung1995acceptability}

% Need to adopt norms so that set Omega stays conflict-free.

% Need to curtail norms when necessary

% Notation of Oren et al. \cite{oren2008argumentation}: $O_{bob}(theatre)$, etc.

% ------------------------------------

In a computerised or non-computerised system, there may exist one or more autonomous entities, which we call agents. In systems containing at least one agent, one may wish to govern agent behaviour either by giving direct commands or by setting guidelines for how the agent should interact with its environment. We call such guidelines \textbf{norms}. If an agent is given more than one norm to follow, some of these norms may contradict one another on how the agent should act in certain situations. We call this a \textbf{normative conflict}. The goal of this paper is to devise a way of resolving normative conflicts by deciding which norms to keep and which norms to modify or remove such that the agent continues to follow the most relevant and most important norms.

A norm may be an obligation, a prohibition, or a permission. As an example, let us consider agent Andy, who works in retail and must follow the given norms:

\begin{adjustwidth}{0.7cm}{0.7cm}
\textbf{Norm 1:} \textit{``Andy is \textbf{obligated} to go to work each weekday"}.\newline 
\textbf{Norm 2:} \textit{``Andy is \textbf{prohibited} from disclosing confidential work information to  employees of other companies"}.\newline 
\textbf{Norm 3:} \textit{``Andy is \textbf{permitted} to take a day off of work if he has requested it in advance"}.
\end{adjustwidth}

Here, we see an example of each type of norm. Generally, a permission is able to override an obligation --- in this case, if Andy were to request a day off of work, then norm 3 would override norm 1. Norms may also be \textit{conditional} --- in norm 3, the condition for activation is that Andy must have requested the day off. 

Prohibitions and obligations can conflict with one another. For example, if Andy's company is working together with another company called CompanyCorp on a project, we may decide to introduce a fourth norm together with the previous three:

\begin{adjustwidth}{0.7cm}{0.7cm}
\textbf{Norm 4:} \textit{``Andy is \textbf{obligated} to share information with CompanyCorp about the joint project"}.
\end{adjustwidth}

Now norms 2 and 4 conflict with one another, so we must decide which norms must be prioritised. Two obligations may also conflict with one another if both cannot be fulfilled simultaneously. Suppose we impose the following new norms:

\begin{adjustwidth}{0.7cm}{0.7cm}
\textbf{Norm 5:} \textit{``Andy is \textbf{obligated} to immediately answer the phone when a customer calls"}.\newline 
\textbf{Norm 6:} \textit{``Andy is \textbf{obligated} to attend to customers at the cash register between 10:00-11:00 each weekday"}.
\end{adjustwidth}

Then if the phone were to ring whilst Andy is attending to a customer at the register, he would be unable to fulfil his obligation to answer the phone. Norms 5 and 6 also show examples of time dependencies within norms. Norm 5 is time-sensitive in its fulfilment condition in that Andy must answer \textit{immediately} (presumably within a given time limit). On the other hand, norm 6 is time-sensitive in that its effects only apply at specific times. For most of this paper, we will not be thoroughly analysing time dependencies of norms.

To resolve normative conflicts, one needs to decide to prioritise certain norms above others, since by definition it is not possible for an agent to follow all conflicting norms it is given. As described by Oren et al., \textit{``An agent may drop a conflicting norm [...] or temporarily ignore it [...] We assume that an agent in this situation has no choice but to drop a norm permanently, and must determine which norms to drop"} \cite{oren2008argumentation}. We will also make this assumption for the initial sections in this paper, but we will also consider situations later where agents can temporarily ignore norms. When we do so, the specifics of how our assumptions change will be discussed. For now, we will assume that certain norms must be dropped whilst others must be adopted (or admitted). Our goal is to adopt a set of norms which do not conflict with one another such that the set contains the most important and most relevant norms possible. This will ensure that the agent(s) following these norms have a clear, well-defined objective which aligns with the tasks of highest priority. This will also ensure that the agent need not engage in normative reasoning whilst it is deployed, allowing its resources to instead be put towards its own agenda.

In this paper, we will introduce a method for deciding which norms to adopt and for resolving normative conflicts in a way which is versatile and coherent. We want this method to be compatible with some common policies for normative conflict resolution (which we will discuss in detail in Section \ref{sectionlitreview}), so this method \textbf{should be adaptable with respect to policies}. This will ensure that the method we introduce will be as flexible as possible to maximise applicability in different settings. We also want the results of this method to \textbf{demonstrably align with notions in argumentation} \cite{dung1995acceptability} --- that is, we want to be able to show that this method produces a set which is admissible or is a equivalent to an extension (these notions will be discussed in section \ref{sectionargsub}). Achieving this goal would ensure that our method produces solutions which are logically sound and which prioritise norms sensibly.

Further to the above goals, we want to \textbf{extend our method to allow for adoption of norms which directly conflict with one another}. In doing so, the set of norms we admit will be moved past the semantics of argumentation, where conflicts can only be resolved by admitting one argument (or norm) or the other.

The structure of the remainder of this paper is as follows:

\begin{itemize}\itemsep0em
    
    \item In Section \ref{sectionargmain}, we give a brief introduction to the existing tools we will use for resolving normative conflicts --- namely argumentation (Section \ref{sectionargsub}), conflict graphs (Section \ref{sectionconflictgraphs}), and graph colouring (Section \ref{sectiongraphcolouring}).
    
    \item In Section \ref{sectionlitreview}, we discuss relevant work on policies which have been used for normative conflict resolution --- these policies give us a specific objective in judging the adaptability of our system. We also discuss relevant work within argumentation.
    
    \item In Section \ref{sectionnewstuff}, we introduce new methods for resolving normative conflicts using the tools discussed in Section \ref{sectionargmain}. In Section \ref{sectionmainalg}, we introduce $ColourResolve$, our first new method which we will build upon throughout the remainder of Section \ref{sectionnewstuff}. Section \ref{sectionheuristics} discusses how our new method can be used with different policies. Section \ref{sectioncompleteextension} highlights a shortcoming of our new result and introduces a remedy, $ColourResolveComplete$. Section \ref{sectioncurtailment} further refines our method, resulting in the most robust of all the new methods introduced so far, $ColourCurtailComplete$ --- this method's computational complexity is discussed in Section \ref{sectioncomplexity}.
    
    \item In Section \ref{sectionresults}, we compare the empirical results of the new methods we have introduced to existing methods in literature using existing methodology. We also perform our own empirical evaluations.
    
    \item In Section \ref{sectionfurtherwork}, we discuss potential areas for further exploration beyond our work.
    
    \item Finally, we conclude in Section \ref{sectionconclusion}.
\end{itemize}

\section{Argumentation and graph colouring} \label{sectionargmain}

In standard logic, the discovery of new information does not alter the conclusions at which one has already arrived. The counterpart to this type of reasoning is called \textit{non-monotonic} logic, where conclusions can be withdrawn or overturned on the basis of new information. For example, if one were looking at a starry sky at night and saw a light, it might be assumed that the light was a star. However, if one then sees that the star is moving, it may be more reasonable to instead assume that the light is a meteor. If one then sees that the light is flashing, one could again revise their assumption and presume that the light belongs to an aeroplane. Non-monotonic reasoning provides a basis for argumentation, which we will now discuss.

\subsection{Argumentation} \label{sectionargsub}

We will need tools to allow us reason about the conflicts between norms directly. To do so, we can use \textbf{argumentation}, a type of framework pioneered by Dung in 1995 \cite{dung1995acceptability}. Argumentation involves a set of \textit{arguments} $A$ and an \textit{attack relation} $R$ between these arguments. For two arguments $a,b \in A$, if there is a tuple $(a,b) \in R$, we say that argument $a$ attacks argument $b$. In our case, we will view each norm as an argument (and will henceforth use ``norm" and ``argument" interchangeably) and we will view a conflict between two norms as their corresponding arguments attacking one another. Note that we will therefore be considering attack arguments to be bidirectional --- if argument $a$ attacks argument $b$, then argument $b$ attacks argument $a$. This is not usually the case in argumentation, but this assumption will simplify the remainder of this paper.

By using argumentation, we now have access to notions within the field which are used to reason about which arguments are reasonable and should be accepted and which arguments should be rejected. This will help us resolve conflicts by discarding norms we value less highly than others when a conflict arises. The following list contains some common notions within argumentation which we will be using throughout the remainder of this paper:

\begin{adjustwidth}{0.7cm}{0.7cm}
\textit{Note:} We will use $\Omega$ to denote a subset of the set all arguments $A$. This is non-standard notation which we are adopting due to a notational conflict which will arise later in Section \ref{sectiongraphcolouring}.
\end{adjustwidth}

\begin{itemize}\itemsep0em
\item We say that a set of arguments $\Omega$ is \textbf{conflict-free} iff there are no arguments $a,b \in \Omega$ s.t.\ $a$ attacks $b$

\item We say that an argument $a\in A$ is \textbf{acceptable} wrt.\ a set of arguments $\Omega$ iff for any $b\in A$ s.t.\ $b$ attacks $a$, there exists some $c\in \Omega$ s.t.\ $c$ attacks $b$. That is, $\Omega$ attacks every argument which attacks $a$ (or in other words, $\Omega$ \textit{defends} $a$).

\item We say that a set of arguments $\Omega$ is \textbf{admissible} iff it is conflict-free and for every $a\in \Omega$, $a$ is acceptable wrt.\ $\Omega$

\item A set of arguments $\Omega \subset A$ is a \textbf{complete extension} of $A$ iff $\Omega$ is admissible and for every $a\in A$ which is acceptable wrt.\ $\Omega$, $a\in \Omega$

\item A set of arguments $\Omega \subset A$ is a \textbf{preferred extension} of $A$ iff $\Omega$ is a \textit{maximal} admissible set. That is, $\left|\Omega\right| = \max \big\{\left|E\right| \big| E\subset A \text{ is admissible wrt.\ } A\big\}$.
\end{itemize}

There are also other notions and types of extensions within argumentation, such as stable extensions and grounded extensions, but we will not consider these in this paper. For further reading on notions in argumentation, the reader is directed to Dung's seminal paper \cite{dung1995acceptability}.

\subsection{Conflict graphs}\label{sectionconflictgraphs}

Though there are various ways of representing normative conflicts, we will be using \textbf{conflict graphs}, put forth by Oren et al.\ \cite{oren2008argumentation}. The advantage of conflict graphs is that we can use existing work from both argumentation and graph theory to our advantage when resolving conflicts, as well as standard normative reasoning.

In standard graph theory, a \textbf{graph} consists of a finite set of discrete vertices $V$ (often also called nodes, as with Oren et al.) and a set of edges $E$. Each edge consists of a pair of two vertices $v,w \in V$ and can be thought of as a link or a relation between vertices $v$ and $w$. Edges may be unordered pairs $\{v,w\}$ or ordered pairs $(v,w)$ --- if a graph contains ordered pairs as edges, then these edges are said to be directional, and the corresponding graph is called a directed graph, or a digraph. Directional edges represent a relation which is one-sided. On the other hand, graphs without directed edges are called undirected graphs, or simply ``graphs".

In a conflict graph, each vertex represents a norm (or an argument), while each edge represents a normative conflict between two norms. We will assume that edges are undirected --- that is, whenever norm $v$ conflicts with norm $w$, norm $w$ also conflicts with norm $v$. A potential area for further work may involve directional conflicts, but we will not consider these cases in this paper.

As an example, let us consider a system where our agent Andy from before has been given the six norms we discussed in Section \ref{sectionintroduction}:

\begin{adjustwidth}{0.7cm}{0.7cm}
\textbf{Norm 1:} \textit{``Andy is \textbf{obligated} to go to work each weekday"}.\newline 
\textbf{Norm 2:} \textit{``Andy is \textbf{prohibited} from disclosing confidential work information to  employees of other companies"}.\newline 
\textbf{Norm 3:} \textit{``Andy is \textbf{permitted} to take a day off of work if he has requested it in advance"}.\newline 
\textbf{Norm 4:} \textit{``Andy is \textbf{obligated} to share information with CompanyCorp about the joint project"}.\newline 
\textbf{Norm 5:} \textit{``Andy is \textbf{obligated} to immediately answer the phone when a customer calls"}.\newline 
\textbf{Norm 6:} \textit{``Andy is \textbf{obligated} to attend to customers at the cash register between 10:00-11:00 each weekday"}.
\end{adjustwidth}

\begin{figure}[!h]
    \centering
    \includesvg[width=0.3\textwidth]{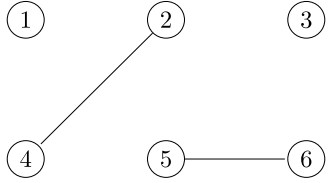}
    
    \caption{A conflict graph representation of the set of norms discussed in Section \ref{sectionintroduction}. The number inside each vertex corresponds to the number of the norm it represents. An edge connecting two vertices represents a conflict between the two corresponding norms.}\label{andyconflictgraph}
\end{figure}

Here, we have that norms 2 and 4 directly conflict with one another in their statements. Whenever Andy is busy attending to customers, there is also a conflict between norms 5 and 6 since Andy would be unable to fulfil both duties at once. Therefore, the normative conflict graph would appear as shown in Figure \ref{andyconflictgraph}. Note that we have not drawn an edge from norm 1 to norm 3 because Andy's permission to take time off of work can override his obligation to go to work when norm 3 is applicable.

\subsection{Graph colouring} \label{sectiongraphcolouring}

% Describe basic graph theory

% Describe graph colouring. NP-complete, but p-time algorithms exist, like DSatur. will discuss later in Section (blah).

% Give an example of a coloured graph

As we continue, we will use \textit{graph colouring} as one of our tools to solve normative conflicts, which we will introduce in this section. Given a graph consisting of vertices and edges, the aim of graph colouring is to assign each vertex a colour such that no two vertices of the same colour are connected by an edge --- typically, one tries to accomplish this using the fewest possible number of colours.

A graph colouring which uses at most $k$ colours is called a \textit{$k$-colouring}, and a graph $G$ is said to be \textit{$k$-colourable} if there exists a proper $k$-colouring for $G$. The minimum $k$ for which a graph $G$ is $k$-colourable is called the \textit{chromatic number} of $G$, denoted $\chi(G)$. For a given colour $c$, the set of vertices corresponding to colour $c$ is called the \textit{colour class} of $c$. 

\begin{figure}[!h]
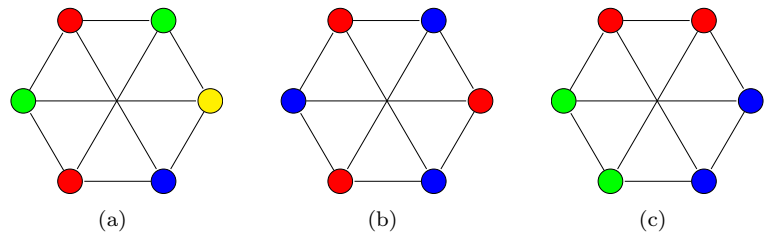

    \centering
    \subfigure[]{\includesvg[width=0.2\textwidth]{images/colouringexample3.svg}}
    \; \; \;
    \subfigure[]{\includesvg[width=0.2\textwidth]{images/colouringexample1.svg}}
    \; \; \;
    \subfigure[]{\includesvg[width=0.2\textwidth]{images/colouringexample4.svg}}
    
    \caption{An example of a graph with --- (a) A valid 5-colouring. (b) A valid 2-colouring. (c) An invalid 3-colouring.}\label{examplescolouring}
\end{figure}

In general, deciding whether there exists a $k$-colouring for $k \in \mathbb{N}\setminus \{0,1,2\}$ for a graph is an NP-complete problem. Furthermore, finding the chromatic number $\chi(G)$ for a graph $G$ is an NP-hard problem \cite{karp1972reducibility}. As such, we will not focus on finding graph colourings which use the exact minimum number of colours. Instead, we will consider known algorithms for creating colourings which may use more colours than the chromatic number, which we will discuss later. Some examples of possible colourings for a graph can be seen in Figure \ref{examplescolouring}. For further reading on graph colouring, the reader is directed to Lewis' book on graph colouring \cite{lewis2015guide}.

Note that for any valid colouring of any graph, each colour class is an \textit{independent set} --- that is, there are no edges connecting any two vertices in the same colour class. This fact will serve as the backbone of the algorithms we will propose, where we will compare colour classes to find independent sets which correspond to the most important norms in the system.

As we continue into subsequent sections, we will use the following notation:

\begin{itemize}
    \item We will use standard graph-theoretic notation for most objects. $G=(V,E)$ will represent a graph with vertex set $V$ and edge set $E$.

    \item When colouring any graph, one can create a bijection from the set of colours to $\{1,2,3,...,n\} \subset \mathbb{N}$, where $n$ is the number of colours being used in the graph (so $n\leq\left|V\right|$). As such, $\mathbb{N}$ will represent the space of all possible colours which can be used and $c\in \mathbb{N}$ will represent an arbitrary colour.
    
    \item For a graph $G=(V,E)$, a graph colouring $\phi$ is equivalent to a function $\phi:V \rightarrow \mathbb{N}$ since each vertex maps to a colour and each colour can be represented by an arbitrary natural number. We will denote arbitrary graph colourings by $\phi$ and the set of possible graph colourings by $\Phi$.
    
    \item Since $E$ represents the edge set of a graph, this notation conflicts with the standard notation for an extension or a set of arguments in argumentation (also $E$). Therefore, we will use $\Omega$ to denote a set of arguments/norms or a selected set of vertices while $E$ will represent the edge set of a graph.
    
    \item We will use $A$ to denote the set of all arguments in the domain of our problem. Since we are modelling norms as arguments and arguments as vertices in a conflict graph, the set of arguments $A$ and the set of vertices $V$ represent the same objects seen through a different level of abstraction.

\end{itemize}

\section{Related work on conflict resolution}\label{sectionlitreview}

Having introduced argumentation, we can now discuss existing materials on normative conflicts. Normative conflicts can be resolved using many different approaches, some of which involve argumentation, while others do not. In Section \ref{sectionrelatednorm}, we discuss how \textit{policies} for normative conflict resolution are used, both in settings involving and not involving argumentation. In Section \ref{sectionrelatedarg}, we discuss the relation between argumentation and normative reasoning, and also discuss some criteria for how arguments be used to can defeat one another, building upon the policies discussed in Section \ref{sectionrelatednorm}.

\subsection{Policies for normative conflict resolution}\label{sectionrelatednorm}

One example of a method for resolving normative conflicts without argumentation comes from Vasconcelos et al.\ \cite{vasconcelos2009normative}. They show that a normative conflict arises when any action is \textit{within scope} of two conflicting norms, such as a prohibition and an obligation. That is, a conflict exists when an action is both simultaneously forbidden and obliged by some agent. Vasconcelos et al.\ also discuss policies for resolving conflicts, including \textit{lex posterior} (where the more recent norms takes priority over less recent norms) and \textit{lex superior} (where norms put forth by greater authorities take priority over others).

An appropriate example for demonstrating another type of policy comes from Horty \cite{horty1994moral}, who put forth the following set of conflicting norms as a motivating example:

\indent \indent \textit{``Don't eat with your fingers,}

\indent \indent \textit{If you are served asparagus, eat it with your fingers."} \cite{horty1994moral}

Here, neither lex superior nor lex posterior are applicable as we do not have information on the sources of norms nor on when they were imposed. Instead, the policy of \textit{lex specialis} (where norms which are more specific than others take priority) is more suitable, since the first norm should be ignored when presented with asparagus. Stra\ss er and Arieli \cite{strasser2019normative} have used an approach based on lex specialis to resolve normative conflicts through argumentation. In their work, arguments (or sequents as they call them) are eliminated when they contain another argument which attacks them either via rebuttal (directly refuting the opposing conclusion) or undercutting (refuting one or more of the opposing antecedents).

One of our previously stated aims for creating a new framework for conflict resolution is that we wish for it to be adaptable with respect to policies. Setting this as a specific objective, \textbf{we wish for our resulting system to be compatible with the policies of lex superior, lex posterior, and lex specialis}. 

Normative conflicts could also be resolved based on a system of trust. This bears a similarity to how one would apply lex superior, though instead of considering the amounts of authority of individuals who impose norms, we instead consider individuals based on their reputation. Examples include the work of Parsons et al.\ \cite{parsons2011using} \cite{parsons2011argumentation} and of Keung et al.\ \cite{keung2008using}. However, trust-based systems require more involved processes than the previously discussed policies since one would have to derive a measure of trust for each individual rather than using easily obtainable information within the system. That said, given a sufficiently adaptable system, should be possible to include trust-based policies as an ``add-on" to the framework.

\subsection{Policies and argumentation} \label{sectionrelatedarg}

Similarly to how policies exist for resolving normative conflicts, there are various criteria which can be used in argumentation for determining which arguments defeat others. One such example is the \textit{weakest link} criterion. 
Consider a system where each argument is composed of several elementary arguments, called context arguments, where the conclusion of each context argument leads to the antecedent of the subsequent context argument. Here, the weakest link criterion dictates that the strength of an argument is determined by the strength its weakest context argument \cite{liao2016prioritized}. Another example would be the \textit{last link} criterion, where the strength of an argument is determined by the strength of its final context argument. Our framework will not consider these criteria for defeating arguments and will instead focus on policies which do not involve argumentation-based reasoning. However, an alternate system based upon ours which uses argumentation-specific criteria to decide which norms to follow may be an avenue for future work.

Other criteria for defeating arguments exist where arguments can be attacked without being defeated. For example, Amgoud and Ben-Naim \cite{amgoud2013ranking} have proposed a system where arguments are sorted by strongest to weakest based upon how many attackers an argument has.

Some frameworks and methods for non-monotonic reasoning have later been shown to have equivalent approaches using argumentation. For example, Liao et al \cite{liao2016prioritized} showed that using a ``greedy" approach for creating an extension using the weakest link criterion creates a set of conclusions identical to those of an approach by Brewka and Eiter \cite{brewka1999preferred} from 17 years prior. It is expected that these parallels between argumentation and non-monotonic logic should exist, since it has been shown that Dung's argumentation framework is equivalent to classical logic with the addition of the Peirce-Quine dagger (i.e. the ``NAND" operator) \cite{gabbay2011dung}. As such, approaches with and without argumentation should not be seen as mutually exclusive, with various degrees of overlap existing between different approaches. Therefore, by applying argumentation to normative conflict resolution, loss of detail or generality should not be considered a major concern.

It should be noted that the approaches featured in \cite{liao2016prioritized} and \cite{brewka1999preferred} require the existence a preference relation over context arguments, which could be derived using policies as in Section \ref{sectionrelatednorm}, or they could be given a priori. Other argumentation-based systems, such as those of Modgil and Luck \cite{modgil2008argumentation}, do not require an a priori relation, while other systems which do not involve argumentation also do not require an a priori relation \cite{vasconcelos2009normative}. We can therefore see that whether or not one chooses to involve argumentation in a system or framework does not directly impose a requirement (or lack thereof) for an ordering to be given as a component of the system. However, it is more advantageous to have this relation be derived as a part of the system. As explained by Modgil and Luck,  \textit{``flexible and adaptive agents need to engage in motivational argumentation over the respective merits of goals. Therefore, argumentation frameworks in which preferences [...] are ‘given’, and not themselves subject to reasoning, do not suffice."} \cite{modgil2008argumentation}. We will follow this philosophy with our own framework, aiming to \textbf{use preference orderings which can be derived from the properties of norms and arguments}, rather than relying on orderings given a priori.

\section{Normative conflict resolution}\label{sectionnewstuff}

\subsection{Graph colouring for normative conflicts}\label{sectionmainalg}

% Describe conflict graphs - Oren et al.

% We will assume that conflict graphs have already been pruned as with Oren et al.

Now that we have access to both graph colouring and conflict graphs, we can use these two tools to begin to solve normative conflicts. Now assume we have an arbitrary \textit{heuristic} function $h: G \times \Phi \times \mathbb{N} \rightarrow \mathbb{R}$ --- calling $h$ would appear in the form $h(G,\phi,c)$. That is, $h$ takes a graph, a colouring, and a colour as inputs and evaluates the overall importance of the set of norms corresponding to the given colour by returning a real number.

Let us now introduce our first new algorithm \textit{ColourResolve} in Algorithm \ref{mainalg}, which uses graph colouring, conflict graphs, and an arbitrary heuristic to resolve normative conflicts.

\begin{algorithm}
\textbf{Inputs:} A conflict graph $G=(V,E)$. A heuristic $h:G\times \Phi\times \mathbb{N} \rightarrow \mathbb{R}$.

Use a graph colouring algorithm of choice to create a $k$-colouring $\phi:V \rightarrow \mathbb{N}$

$c_{\text{best}} \gets \argmax\limits_{c \in \{1,...,k\}} \Big( h(G, \phi, c) \Big)$\label{linea1} %$h(G)$

$V^\prime \gets \{v \in V \mid \phi(v)=c_{\text{best}} \}$ \label{linea2}

$\Omega \gets \{\text{Norm corresponding to } v \mid v \in V^\prime \}$ \label{linea3}

\textbf{Return} $\Omega$ \label{linea4}

\caption{\textit{ColourResolve(G)} --- resolves normative conflicts in a conflict graph through graph colouring.}\label{mainalg}
\end{algorithm}

\textit{ColourResolve} first creates a colouring using an arbitrary graph colouring algorithm (we will discuss this further later). On line \ref{linea1}, it then uses the given heuristic to find the colour which maximises the heuristic value, then assigns this to the variable $c_{\text{best}}$. On line 4, it finds the subset of vertices $V^\prime \subset V$ corresponding to the colour $c_{\text{best}}$. On line \ref{linea3}, the norms corresponding to the vertices with colour $c_\text{best}$ are found, which are then returned on line \ref{linea4}. \textit{ColourResolve} always produces an admissible set, which we will now prove.

\begin{proposition}\label{propadmissible}
Let $G=(V,E)$ be any conflict graph. When using \textit{ColourResolve} on $G$ with any proper graph colouring algorithm and any heuristic $h$, the resulting set of arguments $\Omega$ is always admissible.
\end{proposition}

\begin{Proof}

To show that $\Omega$ is admissible, we need to demonstrate both of the following facts:

\begin{enumerate}\itemsep0em
    \item $\Omega$ is conflict-free.
    \item For each $a\in \Omega$, $a$ is acceptable wrt.\ $\Omega$. That is, for each $b\in A$ s.t. $b$ attacks $a$, there exists $c \in \Omega$ s.t. $c$ attacks $b$.
\end{enumerate}

Let us start with showing that statement 1 holds. Suppose for contradiction that $\Omega$ was not conflict-free. Then there exist two arguments $u,v \in \Omega$ such that $u$ attacks $v$. Therefore, there exists an edge $\{u,v\} \in E$ in the conflict graph. Since $u$ and $v$ are both in $\Omega$, they must have been assigned the same colour regardless of which heuristic and graph colouring algorithm was used. However, it is not possible to assign the same colour to two vertices connected by an edge in a graph colouring problem. Thus, a conflict cannot exist and so $\Omega$ must be conflict-free.

As for statement 2, let $a \in \Omega$ and let $b \in A$ s.t. $b$ attacks $a$. Then there is a normative conflict between $a$ and $b$, so there is a bidirectional edge $\{a,b\}$ in the conflict graph $G$. Therefore $a$ attacks $b$ and $a\in \Omega$, so $a$ defends itself.

Since statements 1 and 2 hold, we can conclude that $\Omega$ is admissible.
\end{Proof}

Since both the graph colouring algorithm and heuristic in Algorithm \ref{mainalg} are arbitrary, \textit{ColourResolve} is flexible and can be adapted to suit the needs of the situation at hand. For example, one could use advanced graph colouring algorithms to minimise the number of colours used in the graph, which would thereby maximise the number of vertices corresponding to any given colour --- this would therefore maximise the number of norms which are admitted at the end of the algorithm. Alternatively, one could use simpler graph colouring algorithms to save computational time. As an example of a polynomial time algorithm one could use, we will briefly outline the \textit{DSatur} algorithm \cite{brelaz1979new}, which has worst-case running time $\mathcal{O}(n^2)$ \cite{lewis2015guide}. This was first put forth by Br\'elaz in 1979 \cite{brelaz1979new} and can be seen in Algorithm \ref{dsaturalg}.

\begin{algorithm}
\textbf{Input:} A graph $G=(V,E)$

\While{not all vertices have been coloured}{
    Select an uncoloured vertex with maximal degree of saturation. Break ties by selecting the vertex of highest degree. Break further ties arbitrarily.
    
    \For{each colour $c$ used so far}{
        Attempt to assign colour $c$ to vertex $v$
        }
    \If{no colours were suitable}{
        Assign $v$ to a new colour
    }
}

\caption{DSatur graph colouring algorithm by Br\'elaz \cite{brelaz1979new}}\label{dsaturalg}
\end{algorithm}

Here, the \textit{degree of saturation} of a vertex $v$ refers to the number of coloured vertices $v^\prime$ with an edge $(v,v^\prime)$ or $(v^\prime,v)$. DSatur prioritises colouring vertices with the highest degree of saturation, thereby focusing on assigning colours to vertices which have the fewest colour options available. Ties are broken by selecting the highest-degree vertex, allowing DSatur to focus on vertices which impose the most constraints upon other vertices.
The intended purpose of this example is to show the reader a powerful yet intuitive graph colouring method.

\subsection{Heuristics}\label{sectionheuristics}

To demonstrate the flexibility of using arbitrary heuristics, we will now define some possible heuristics one could use with \textit{ColourResolve}. Definition \ref{deflexpos} defines a version of \textit{lex posterior} adapted for use as a heuristic and Definition \ref{deflexsup} defines an adapted version of \textit{lex superior}. Definition \ref{defhnum} defines a heuristic which selects the colour class corresponding to the greatest number of vertices and therefore admits the greatest number of norms.

\begin{definition}\label{deflexpos} Lex posterior (heuristic version)

Given a graph $G=(V,E)$, a colouring $\phi:V \rightarrow \mathbb{N}$, and a colour $c \in \mathbb{N}$, define the heuristic $h_\text{pos}:G\times\Phi\times\mathbb{N} \rightarrow \mathbb{R}$ by:

\begin{equation*}
h_{\text{pos}}(G, \phi, c) \coloneqq  \sum_{v \in V \land \phi(v)=c} \left|\Big\{v^\prime\big|\{v,v^\prime\} \in E , T_v < T_{v^\prime}\Big\} \right| \end{equation*}

Here, $T_v$ refers to the \textit{time} at which the norm corresponding to vertex $v$ was imposed.
\end{definition}

The heuristic $h_\text{pos}$ finds the number of times any argument corresponding to a vertex with colour $c$ defeats another argument by being declared earlier.

Our definition of $h_\text{pos}$ checks for an edge ``$\{v,v^\prime\} \in E$". If we were to generalise our methods to directed graphs, then this edge could either be $(v,v^\prime)$ or $(v^\prime,v)$, where we would accept either option. We do not need to check the colour of each vertex $v^\prime$ which each $v$ attacks since no two vertices with the same colour can be connected by an edge. 

One could further modify this definition by subtracting the number of times each argument corresponding to a vertex with colour $c$ is defeated by another argument. In section \ref{sectionresults} later, we will perform empirical evaluations on our newly proposed algorithms. There, we will use the modified version of this definition (and similarly modified versions of other definitions in this section). However, the version shown in Definition \ref{deflexpos} is simpler to state, which is why this definition has been declared in this way.

%h_{\text{pos}}(G, \phi, c) \coloneqq \left. \Bigg\{ \; \left|  \Big\{v^\prime\big|(v,v^\prime) \in E, T_v < T_{v^\prime}\Big\}    \right|    \; \right| \;  v \in V, \phi(v)=c \; \Bigg\} \end{equation*} % The sum of the amount of neighbours each node defeats under lex posterior

\begin{definition}\label{deflexsup} Lex superior (heuristic version)

Given a graph $G=(V,E)$, a colouring $\phi:V \rightarrow \mathbb{N}$, and a colour $c \in \mathbb{N}$, define the heuristic $h_\text{sup}:G\times\Phi\times\mathbb{N} \rightarrow \mathbb{R}$ by:

\begin{equation*}
h_{\text{sup}}(G, \phi, c) \coloneqq  \sum_{v \in V \land \phi(v)=c} \left|\Big\{v^\prime\big|\{v,v^\prime\} \in E , \rho_v \succ_\rho \rho_{v^\prime}\Big\} \right| \end{equation*}

Here, $\rho_v$ refers to the power of the authority who imposed the norm corresponding to vertex $v$. 
\end{definition}

The preference relation $\succ_\rho$ is defined by a weak ordering defined as follows: 

\begin{center}
For all norms $a,b$: $\rho_a \succ_\rho \rho_b \Leftrightarrow a$ was imposed by a stronger power than $b$
\end{center}

This relation is a simplified version of one used by Vasconcelos et al.\ \cite[Section 7]{vasconcelos2009normative}. Again, as with our definition for lex posterior, we check for edges going in both directions --- $(v,v^\prime)$ and $(v^\prime, v)$. We can also modify the definition by subtracting the number of times each argument with the specified colour is defeated.

In fact, this preference relation can be generalised to fit any preference ordering over vertices in a conflict graph, and new policies can be created when and where necessary. Definition \ref{defgeneral} gives a generalised version of Definitions \ref{deflexpos} and \ref{deflexsup} which is compatible with any weak ordering over all vertices in the conflict graph.

\begin{definition}\label{defgeneral} Heuristic for any general preference relation (generalises definitions \ref{deflexpos} and \ref{deflexsup})

Given a graph $G=(V,E)$, a colouring $\phi:V \rightarrow \mathbb{N}$, and a colour $c \in \mathbb{N}$, and some weak ordering $\succ$ on $V$, define the heuristic $h_\succ:G\times\Phi\times\mathbb{N} \rightarrow \mathbb{R}$ by:

\begin{equation*}
h_\succ (G, \phi, c) \coloneqq  \sum_{v \in V \land \phi(v)=c} \left|\Big\{v^\prime\big|\{v,v^\prime\} \in E , v \succ v^\prime\Big\} \right| \end{equation*}

\end{definition}

We see that Definition \ref{defgeneral} becomes equivalent to Definition \ref{deflexpos} if the given weak ordering depends only the time at which norms were declared. It also becomes equivalent to Definition \ref{deflexsup} if the given weak ordering depends only on the power of the authority who imposed the given norm. We can also see that Definition \ref{defgeneral} generalises \textit{lex specialis} if the given weak ordering is defined by the following:

\begin{center}
    For all norms $a,b$: $a \succ b \Leftrightarrow$ ant($a$) $\subsetneq$ ant($b$)
\end{center}

...where ant($x$) is the set of antecedents of $x$ for any argument $x$ (or in the case of a norm, ant($x$) corresponds to the requirements for norm $x$ to come into effect).

Since we have shown that $ColourResolve$ is compatible with lex posterior, lex superior, and lex specialis, we have met our criteria for showing that our system is adaptable with respect to conflict resolution policies. In later sections, we will be modifying $ColourResolve$, but as long as the system continues to use the notion of heuristics, it will remain policy-adaptable.

There is one more heuristic of note which we will define before we consider it further in Section \ref{sectioncompleteextension}. This heuristic selects the colour class which corresponds to the greatest number of vertices overall, and can be seen in Definition \ref{defhnum}.

\begin{definition}\label{defhnum} Maximal colour class heuristic

Given a graph $G=(V,E)$, a colouring $\phi:V \rightarrow \mathbb{N}$, and a colour $c \in \mathbb{N}$, define the heuristic $h_{\max}:G\times\Phi\times\mathbb{N} \rightarrow \mathbb{R}$ by:

\begin{equation*}
h_{\max}(G,\phi,c) \coloneqq  \Big\lvert\big\{ v \in V \mid \phi(v)=c \big\}\Big\rvert \end{equation*}

%That is, $h$ is the heuristic which selects the colour corresponding to the greatest number of vertices in $G$.
\end{definition}

\subsection{Producing argumentation extensions with \textit{ColourResolve}}\label{sectioncompleteextension}

Upon inspection, it may appear that using the $h_{\max}$ heuristic with a suitable graph colouring algorithm for \textit{ColourResolve} would produce a preferred extension. Intuitively, both the preferred extension and the $h_{\max}$ heuristic aim to maximise the number of arguments which are admitted. This certainly is the case with the graph and the colouring shown in Figure \ref{examplescolouring}(b), where the colouring is both optimal and corresponds to a preferred extension under $h_{\max}$. However, although $h_{\max}$ has the potential to \textit{often} produce a preferred extension, this is not always the case.

\begin{figure}[!h]
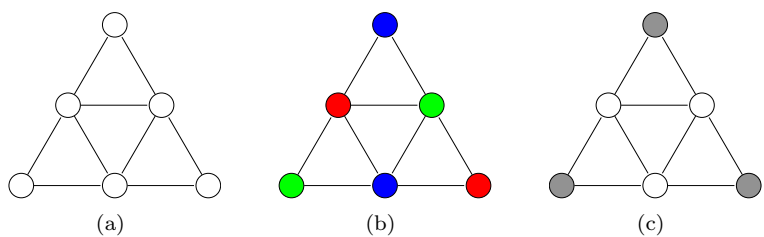

    \centering
    \subfigure[]{\includesvg[width=0.2\textwidth]{images/proofcounterexample2.svg}}
    \; \; \;
    \subfigure[]{\includesvg[width=0.2\textwidth]{images/proofcounterexample.svg}}
    \; \; \;
    \subfigure[]{\includesvg[width=0.2\textwidth]{images/proofcounterexample3.svg}}
    
    \caption{(a) An uncoloured graph. (b) The same graph with a valid 3-colouring. (c) The same graph with an independent vertex set of size 3 (shown in grey).}\label{proofcounterexample}
\end{figure}

Consider the graph shown in Figure \ref{proofcounterexample}(a) -- let us arbitrarily call this graph $G_6$ since it has 6 vertices. It is possible to create a valid 3-colouring for $G_6$, which can be seen in Figure \ref{proofcounterexample}(b). Furthermore, it is impossible to colour $G_6$ using only two colours since $G_6$ contains a fully-connected subgraph with three vertices. In other words, $G_6$ contains ``triangles", and these cannot be coloured with two colours since at least one vertex of the triangle would neighbour a vertex with a colour which matches its own. Therefore, the 3-colouring shown in Figure \ref{proofcounterexample}(b) uses the optimal number of colours \footnote{Here, we say that $G_6$ is 3-chromatic and that $\chi(G_6)=3$.}. By using \textit{ColourResolve} on $G_6$ with heuristic $h_{\max}$ and an exact graph colouring algorithm, we would admit exactly two arguments --- those corresponding to either red, green, or blue.

% https://math.stackexchange.com/questions/684077/graph-colouring-and-maximal-independent-set  ------ graph colouring and maximal independent set
% https://cs.stackexchange.com/questions/109188/connection-between-max-independent-set-and-graph-coloring

If we now consider Figure \ref{proofcounterexample}(c), we can see that $G_6$ contains a vertex subset of three disjoint vertices, shown in grey. If we consider $G_6$ to be a conflict graph, then these grey vertices would correspond to a preferred extension of the corresponding set of arguments. If we wished to colour the graph in Figure \ref{proofcounterexample}, we would need one colour for the three grey vertices, then another distinct colour for each of the three remaining vertices. This would give a 4-colouring, which is sub optimal.

Therefore, an exact graph colouring algorithm used with $h_{\max}$ on a problem with $G_6$ as its conflict graph would admit two arguments, but a preferred extension would contain at least three arguments. So \textit{ColourResolve} does not always find a preferred extension when using an exact colouring.

This example serves to show that although \textit{ColourResolve} is highly versatile, there are situations where the results it produces differ from those of preferred semantics. The caveat here is that we assume that using an exact graph colouring algorithm will allow us to admit the highest possible amount of arguments with heuristic $h_{\max}$, but this may not always be the case. For example, it may be possible that other sub-optimal graph colouring algorithms would assign all vertices in the independent set in Figure \ref{proofcounterexample}(c) to a single colour while assigning each of the remaining vertices their own unique colours. This would create a 4-colouring of $G_6$ and would allow heuristic $h_{\max}$ to admit three vertices despite the colouring being sub-optimal. We can therefore see that colouring corresponding to other types of extensions \textit{exist}, but to find them would depend on the colouring algorithm one uses.

Although a preferred extension is not guaranteed with \textit{ColourResolve}, it is straightforward to guarantee a complete extension by modifying our algorithm. To do so, we will now introduce a variant of \textit{ColourResolve} called \textit{ColourResolveComplete}, which can be seen in Algorithm \ref{variantcompletealg}.

\begin{algorithm}
\textbf{Inputs:} A conflict graph $G=(V,E)$. A heuristic $h:G\times \Phi\times \mathbb{N} \rightarrow \mathbb{R}$.

Use a graph colouring algorithm of choice to create a $k$-colouring $\phi:V \rightarrow \mathbb{N}$

$c_{\text{best}} \gets \argmax\limits_{c \in \{1,...,k\}} \Big( h(G, \phi, c) \Big)$

\textcolor{blue}{\For{each vertex $v \in V$}{\label{linevariant1}
    \If{no neighbouring vertices of $v$ have colour $c_{\text{best}}$}{\label{linevariant2}
        $\phi(v) \gets c_{\text{best}}$\label{linevariant3}}
    }\label{linevariantend}}

$V^\prime \gets \{v \in V \mid \phi(v)=c_{\text{best}} \}$

$\Omega \gets \{\text{Norm corresponding to } v \mid v \in V^\prime \}$

\textbf{Return} $\Omega$

\caption{\textit{ColourResolveComplete(G)} --- a variant of \textit{ColourResolve} which always produces a complete extension. \textcolor{blue}{Lines in blue} are newly added and were not present in the original \textit{ColourResolve}.}\label{variantcompletealg}
\end{algorithm}

The difference between our new algorithm and our original algorithm is the addition of lines \ref{linevariant1}-\ref{linevariantend}. Here, we loop through each vertex in the conflict graph. For each vertex $v$, we check whether there exists a neighbouring vertex, say $w$, such that $\{v,w\}\in E$ and $\phi(w)=c_{\text{best}}$. If no such $w$ exists, then we replace the colour of $v$ with the colour $c_{\text{best}}$. Note that our colouring $\phi$ will always remain valid since we have checked for conflicts between neighbouring vertices before changing any colours. We will now prove that \textit{ColourResolveComplete} always produces a complete extension.

\begin{proposition}\label{propcomplete}
    Given any conflict graph $G=(V,E)$, any heuristic $h$, and any proper graph colouring algorithm, the resulting set $\Omega$ produced by \textit{ColourResolveComplete} will always be a complete extension of $G$.
\end{proposition}

\begin{Proof}
    To show that $\Omega$ is a complete extension, we must demonstrate both of the following:
    
    \begin{enumerate}\itemsep0em
        \item $\Omega$ is admissible.
        \item For every argument $a \in A$ which is acceptable wrt.\ $\Omega$, it holds that $a\in \Omega$.
    \end{enumerate}
    
    For statement 1, the fact that $\Omega$ is admissible follows the same logic as in Proposition \ref{propadmissible}.
    
    For statement 2, let us take any argument $a \in A$ which is acceptable wrt.\ $\Omega$. That is, for any $b \in A$ where $b$ attacks $a$, there exists some $c \in \Omega$ s.t.\ $c$ attacks $b$. We need to show that $a \in \Omega$.
    
    Our algorithm \textit{ColourResolveComplete} created some colouring $\phi$ and chose all vertices corresponding to some colour $c_{\text{best}}$ to be assigned to set $\Omega$. Consider any argument $b \in A$ which attacks argument $a$ --- by assumption, we have that there exists some argument $c \in \Omega$ which attacks argument $b$. Therefore, there exists an edge $\{c,b\}\in E$ in our conflict graph. Since $\phi(c)=c_{\text{best}}$, we can conclude that $\phi(b) \neq c_{\text{best}}$.
    
    Since $b$ attacks $a$, there also exists an edge $\{b,a\}\in E$ in our conflict graph. However, $b$ is any arbitrary argument, so no neighbouring vertices of vertex $a$ have the colour $c_{\text{best}}$ (since $\phi(b) \neq c_{\text{best}}$). Thus, when \textit{ColourResolveComplete} (Algorithm \ref{variantcompletealg}) reaches vertex $a$ on line \ref{linevariant1}, the ``if" statement on line \ref{linevariant2} will resolve to \textit{True}. Therefore $a$ will be assigned colour $c_{\text{best}}$ on line \ref{linevariant3}. 
    
    Hence $\phi(a)=c_{\text{best}}$, so we have that $a \in \Omega$, so statement 2 holds and therefore $\Omega$ is a complete extension.
\end{Proof}

The proof of Proposition \ref{propcomplete} is independent of the order in which we consider vertices to be recoloured, so we will obtain a complete extension regardless of this order. One could iterate through verices in a random order, create an ordering using the heuristic function, or through any other method. However, the complete extension created as a result may be different depending on the method used. Additionally, we cannot recolour all vertices simultaneously since recolouring one vertex may cause another to become the neighbour of a vertex with colour $c_{\text{best}}$. %\footnote{In terms of set theory, the axiom of choice cannot be applied here.}.

\subsection{Admitting further norms through curtailment}\label{sectioncurtailment}

So far, we have resolved normative conflicts by completely omitting one of the two conflicting norms. As such, only a single colour in the colourings we create is of any value to us in creating the output of the algorithms we have used so far. In this section, we will admit norms corresponding to multiple colours into our algorithm's output, making the advantages of involving graph colouring more apparent. 

We have so far used our heuristic function to find a \textit{best} colour, but we can easily extend this to find an entire preference relation over all colours. We would define this preference relation for two colours $c_1,c_2 \in \mathbb{N}$ as $c_1 \succeq c_2$ iff $h(c_1) \geq h(c_2)$. From this, for any $k$-colouring, we can infer a weak ordering of colours $(c_1,c_2,...,c_k)$ where $c_1 \succeq c_2 \succeq ... \succeq c_k$.

With this ordering, we can first admit all arguments corresponding to our most preferred colour, then those corresponding to the next most preferred colour, and so on, until we have admitted as many arguments as possible. However, a key issue arises here in that we cannot admit two conflicting arguments with our current set of tools --- this is where we will need \textit{curtailment}.

Our inspiration for curtailment comes from Vasconcelos et al.\ \cite[Definition 10]{vasconcelos2009normative}. In their work, curtailment is defined rigorously for constraining one norm with respect to another. Their framework takes into account the times at which norms were declared, when norms come into effect, and when norms expire, and they therefore introduce a robust definition of curtailment which utilises these factors. However, we will consider curtailment to be a generalised concept to keep our framework as broadly applicable and as abstract as possible. Given two conflicting norms, norm 1 and norm 2, we will consider the curtailed variant of norm 2 with respect to norm 1 to be as follows:

\begin{adjustwidth}{2cm}{2cm}
\textit{Whenever norm 1 is applicable and it is possible to abide by norm 1, then ignore norm 2. Else, abide by norm 2.}%Abide by norm 1 when it is possible to do so and when norm 1 is in effect. Else, abide by norm 2 instead.}
\end{adjustwidth}

Replacing a norm with its curtailed version eliminates the conflict between the two corresponding norms and therefore removes an edge from the conflict graph. The concept of curtailment can also be viewed as another form of \textit{contrary-to-duty} obligations, where an agent must abide by a secondary norm whenever some primary norm cannot be followed.

\begin{figure}[!h]
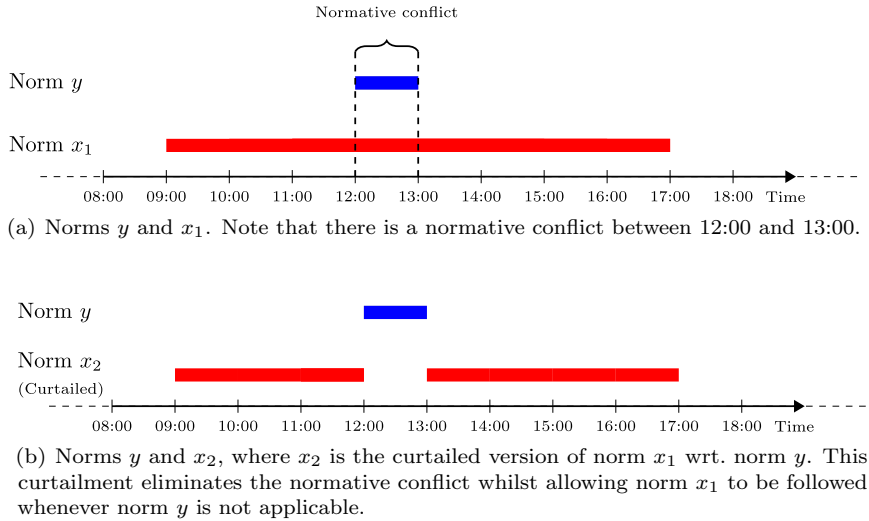

    \centering
    \; \; \;\subfigure[Norms $y$ and $x_1$. Note that there is a normative conflict between 12:00 and 13:00.]{\includesvg[width=0.8\textwidth]{images/curtailexample1crop.svg}}
    \newline\newline\newline
    \subfigure[Norms $y$ and $x_2$, where $x_2$ is the curtailed version of norm $x_1$ wrt.\ norm $y$. This curtailment eliminates the normative conflict whilst allowing norm $x_1$ to be followed whenever norm $y$ is not applicable.]{\includesvg[width=0.8\textwidth]{images/curtailexample2crop.svg}}
    \caption{A timeline showing which norms an agent should follow over time. In (a), we have norm $x_1$ which should be followed between 09:00 and 17:00, and a conflicting norm $y$ which should be followed between 12:00 and 13:00. Norm $x_2$ in (b) is the curtailed version of norm $x_1$ wrt.\ norm $y$, which does not conflict with $y.$}
    \label{figurecurtail}
\end{figure}

As an example, let us consider two norms with time dependencies. Norm $x_1$ says \textit{``Between 09:00 and 17:00, Agent 1 should focus on work tasks"}. Norm $y$ says \textit{``Between 12:00 and 13:00, Agent 1 should take a lunch break"}. By curtailing norm $x$ wrt.\ norm $y$, we obtain a new norm $x_2$ which does not conflict with $y$. Here, $x_2$ would say \textit{``Between 09:00 and 12:00, and between 13:00 and 17:00, Agent 1 should focus on work tasks"}. Figure \ref{figurecurtail}(a) shows norms $y$ and $x_1$ in a timeline, where we can see a conflict between 12:00 and 13:00. Figure \ref{figurecurtail}(b) shows norms $y$ and $x_2$, where the normative conflict has been resolved through curtailment. This allows the agent to follow norm $y$ (take a lunch break) when it is applicable whilst also following norm $x_1$ at all other times. One could view this example as an application of \textit{lex specialis} since we have prioritised norm $y$ over norm $x_1$ as it specifies a narrower time interval.

With curtailment in place, we are now ready to introduce an algorithm which uses every colour in the conflict graph. This can be seen in Algorithm \ref{curtailalg}, which shows \textit{ColourCurtail}.

\begin{algorithm}
\textbf{Inputs:} A conflict graph $G=(V,E)$. A heuristic $h:G\times\Phi\times\mathbb{N} \rightarrow \mathbb{R}$. %A $k$-colouring $\phi$ and set $\Omega$ of admitted arguments or norms (from Algorithm \ref{mainalg}). 

Use a graph colouring algorithm of choice to create a $k$-colouring $\phi:V \rightarrow \mathbb{N}$ \label{lineb1}

$\Omega \gets \varnothing$

\For{$i=1$ to $k$}{\label{lineb2}
    $c_i \gets$ The colour corresponding to the $i^{th}$-highest heuristic value. Break ties arbitrarily. \label{lineb3}

    $V^\prime \gets \{v \in V \mid \phi(v)=c_i \}$ \label{lineb4}
    
    \For{each vertex $v \in V^\prime$}{\label{lineb5}
        \For{%each edge $e \in E$ s.t. $e=(v,w)$ or $e=(w,v)$ for some $w\in \Omega$
        each $w \in \Omega$ s.t. $\{v,w\}\in E$}{\label{lineb6}
            \textbf{CURTAIL} $v$ in-place wrt.\ $w$ \label{lineb7}
            }
        }

    $\Omega \gets \Omega \cup \{\text{Norm corresponding to } v \mid v \in V^\prime \}$}\label{lineb8}

\textbf{Return} $\Omega$

\caption{\textit{ColourCurtail(G)} --- An extended version of \textit{ColourResolve} which uses curtailment to admit every norm.}\label{curtailalg}
\end{algorithm}

The algorithm starts in the same way as \textit{ColourResolve} and \textit{ColourResolveComplete} by creating a $k$-colouring on line \ref{lineb1}. However, rather than admitting only the vertices corresponding to a specified colour, we iterate over colours on line \ref{lineb2}. Line \ref{lineb3} tells us that we iterate over colours in descending order of value according to our heuristic. On line \ref{lineb4} we select the set $V^\prime \subset V$ of vertices corresponding to the colour on the current iteration of the main loop.

On lines \ref{lineb5} and \ref{lineb6}, for each vertex $v$ with the colour in question, we find each vertex $w$ with an edge connected to $v$ where $w$ is already admitted into the final set $\Omega$. Since $w$ is already admitted and $w$ is attacking $v$ or vice versa, either $w$ or $v$ must be \textit{curtailed} with respect to the other. Since $w$ corresponds to a more valuable colour than $v$, we should curtail $v$ with respect to $w$, which we do on line \ref{lineb8}. Since all arguments corresponding to the current colour are curtailed with respect to all arguments admitted so far, we can now admit all arguments which correspond to the current colour, which we do on line \ref{lineb8}. Each time new arguments are admitted, the set $\Omega$ remains conflict-free since each possible conflict has been considered beforehand.

With this algorithm, \textit{every} argument is eventually admitted into the final set, but some may be completely curtailed to the point of having no effect. For example, if we consider our previous example in Figure \ref{figurecurtail}, we curtailed norm $x_1$ wrt.\ norm $y$. However, if we were to instead curtail norm $y$ wrt.\ norm $x_1$, then norm $y$ would no longer have any effect. However, \textit{ColourCurtail} would seek to minimise situations where a higher priority norm is curtailed wrt.\ a lower priority norm since colour classes are considered in descending order of preference according to our heuristic.

As we did with \textit{ColourResolve}, we can make \textit{ColourCurtail} more robust by ``completing" each colour class when we consider it. This results in our final algorithm, \textit{ColourCurtailComplete}, which can be seen in Algorithm \ref{curtailcompletealg}. This utilises all techniques we have seen so far by creating a colouring, iterating through colour classes in descending order of preference, then admitting each colour class with curtailment. The difference between \textit{ColourCurtailComplete} and \textit{ColourResolveComplete} is that when we try to complete a colour class with \textit{ColourCurtailComplete}, we only consider vertices which do not correspond to a colour which has already been admitted into the final set.

On the first iteration of \textit{ColourCurtailComplete}, a set equivalent to \textit{ColourResolveComplete} is created, meaning that this set is a complete extension according to Proposition \ref{propcomplete}. However, on subsequent iterations, we continue to admit further norms into the set. Therefore the set created by \textit{ColourCurtailComplete} does not fit into notions defined within argumentation (since we have directly altered the norms/arguments by curtailing them), but is nonetheless a \textbf{superset of a complete extension}.

\begin{algorithm}
\textbf{Inputs:} A conflict graph $G=(V,E)$. A heuristic $h:G\times\Phi\times\mathbb{N} \rightarrow \mathbb{R}$. %A $k$-colouring $\phi$ and set $\Omega$ of admitted arguments or norms (from Algorithm \ref{mainalg}). 

Use a graph colouring algorithm of choice to create a $k$-colouring $\phi:V \rightarrow \mathbb{N}$

$\Omega \gets \varnothing$

\For{$i=1$ to $k$}{\label{linec1}
    $c_i \gets$ The colour corresponding to the $i^{th}$-highest heuristic value. Break ties arbitrarily.

    \textcolor{blue}{\For{each vertex $v \in V$ which has not been admitted into $\Omega$}{\label{linec2}
        \If{no neighbouring vertices of $v$ have colour $c_i$}{
            $\phi(v) \gets c_i$
        }
    }}

    $V^\prime \gets \{v \in V \mid \phi(v)=c_i \}$
    
    \For{each vertex $v \in V^\prime$}{\label{linec3}
        \For{%each edge $e \in E$ s.t. $e=(v,w)$ or $e=(w,v)$ for some $w\in \Omega$
        each $w \in \Omega$ s.t. $\{v,w\}\in E$}{\label{linec4}
            \textbf{CURTAIL} $v$ in-place wrt.\ $w$
            }
        }

    $\Omega \gets \Omega \cup \{\text{Norm corresponding to } v \mid v \in V^\prime \}$}
    
\textbf{Return} $\Omega$

\caption{\textit{ColourCurtailComplete(G)}  --- a variant of \textit{ColourCurtail} which attempts to ``complete" the colour class upon each iteration. \textcolor{blue}{Lines in blue} are newly added and were not present in the original \textit{ColourComplete}.}\label{curtailcompletealg}
\end{algorithm}

\subsubsection{A note on the relative performances of algorithms introduced so far}\label{sectionnotecomparison}

So far, we have introduced algorithms $Colour\allowbreak Resolve$, $Colour\allowbreak Resolve\allowbreak Complete$, $Colour\allowbreak Curtail$, and $Colour\allowbreak Curtail\allowbreak Complete$. Of these, $ColourResolve$ is our most basic algorithm and serves as a baseline for the remaining algorithms. Given any heuristic, $Colour\allowbreak Resolve\allowbreak Complete$ admits the same number or more norms than $Colour\allowbreak Resolve$ since it ``completes" the extension.

$Colour\allowbreak Curtail$ also outperforms $Colour\allowbreak Resolve$ in terms of norms admitted since it iterates over each colour class to admit curtailed versions of the corresponding norms. On the first iteration of $ColourCurtail$, the norms corresponding to the strongest colour class are admitted, which is equivalent to the result of $ColourResolve$ --- further norms are then admitted on subsequent iterations. We can use similar reasoning to conclude that $ColourCurtailComplete$ outperforms $ColourResolveComplete$

$ColourCurtailComplete$ outperforms $ColourCurtail$ since it ``completes" the colour class at each iteration. Therefore, if a norm is admitted at an earlier iteration (due to completion) than it otherwise would have been, then it will be curtailed with respect to fewer norms than otherwise. Therefore, although $Colour\allowbreak Curtail$ and $Colour\allowbreak Curtail\allowbreak Complete$ both admit all norms, we can see that $Colour\allowbreak Curtail\allowbreak Complete$ admits norms in a less curtailed state.

Since $Colour\allowbreak Curtail\allowbreak Complete$ outperforms $Colour\allowbreak Curtail$ and $Colour\allowbreak Curtail$ outperforms $Colour\allowbreak Resolve$, we see that $Colour\allowbreak Curtail\allowbreak Complete$ also outperforms $Colour\allowbreak Resolve$ by transitivity.

\begin{figure}[!h]
\centering{
    \includesvg[width=0.4\textwidth,inkscapelatex=false]{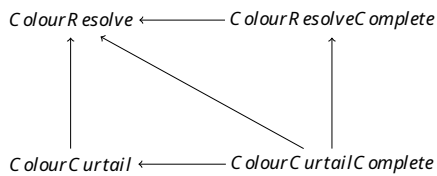}}
    \caption{A diagram comparing the relative performances of algorithms proposed in Section \ref{sectionnewstuff}. An arrow from any algorithm $a$ to any algorithm $b$ represents that algorithm $a$ produces an equal or superior output to algorithm $b$.}\label{diagramalgorithmsperformance}
\end{figure}

A visual summary of which algorithms outperform one another can be seen in Figure \ref{diagramalgorithmsperformance}.

\subsection{Complexity of final algorithm}\label{sectioncomplexity}

Since \textit{ColourCurtailComplete} in Algorithm \ref{curtailcompletealg} is the most robust algorithm out of those we have introduced, let us informally derive its computational complexity.

Suppose we use a graph $G=(V,E)$ as an input into \textit{ColourCurtailComplete}, where $\left|V\right|=n$ for some $n\in \mathbb{N}$. Then in the worst-case scenario, we would use one colour for each vertex, creating an $n$-colouring. This would mean that the loop on line \ref{curtailcompletealg} would be executed $n$ times. Within this loop, on line \ref{linec2} we would consider at most $n$ vertices. However, the other inner loop on line \ref{linec3} is more computationally expensive. Here, we would consider at most $n$ vertices, then for each of these vertices we would consider at most another $n$ vertices (which would happen when the vertex in question is connected to every other vertex via an edge).

Therefore, the worst-case computational \textbf{complexity of \textit{ColourCurtailComplete}} is dependent on three nested loops, each of which considers at most $n$ items, so the complexity is $\mathcal{O}(n\times n\times n)=\pmb{\mathcal{O}(n^3)}$. This result assumes that we are not constrained by the complexity of our graph colouring algorithm (as would be the case with DSatur, for example, which has worst-case running time $\mathcal{O}(n^2)$).

\section{Empirical Evaluation}\label{sectionresults}

So far, we have mathematically proven results about the algorithms we presented, but we yet to empirically assess these algorithms. As $ColourResolve$ and $ColourResolveComplete$ do not involve curtailment and instead consider binary decisions as to whether each norm should be admitted, we can compare these to the results of Oren et al.\ \cite{oren2008argumentation} --- this would be an especially prudent comparison considering that their work formed a basis for our own.

After they presented normative conflict graphs, another contribution from Oren et al.\ is that they found a preferred extension, then admitted each norm corresponding to an argument in the preferred extension. We will compare the performance of $ColourResolve$ and $ColourResolveComplete$ to the performance of that of a preferred extension. However, we do not expect these algorithms to surpass the performance of a preferred extension. As discussed in Section \ref{sectioncompleteextension}, $Colour\allowbreak Resolve$ and $Colour\allowbreak Resolve\allowbreak Complete$ are not guaranteed to find a preferred extension of a graph --- instead, they find an independent set (in the case of $Colour\allowbreak Resolve$) and extend it to a complete extension (in the case of $Colour\allowbreak Resolve\allowbreak Complete$). Both of these notions are subsets (or equal to) a preferred extension. However, the merits of $ColourResolve$ and $ColourResolveComplete$ are that they provide a stepping stone towards creating the more powerful variants, $Colour\allowbreak Curtail$ and $Colour\allowbreak Curtail\allowbreak Complete$. As such, the fact that $Colour\allowbreak Resolve$ and $Colour\allowbreak Resolve\allowbreak Complete$ do not outperform a preferred extension should not diminish their worth. These empirical observations instead serve to provide insight into how much performance is lost by using $ColourResolve$ and $ColourResolveComplete$ rather finding a preferred extension. The smaller the gap between the size of sets of admitted norms created by our algorithms and those of a complete extension, the greater a success we can consider our algorithms to be.

In Section \ref{sectiontest1}, we will briefly summarise evaluations performed by Oren et al.\ on their own contributions. In Section
\ref{sectiontest2}, we will follow the methodology of Oren et al.\ to obtain a direct comparison of our contributions to theirs. We will then perform further evaluations through our own methodology in Section \ref{sectiontest3}.

\subsection{Evaluation by Oren et al.\ on their own work}\label{sectiontest1}

Oren et al.\ \cite{oren2008argumentation} evaluated the performances of three heuristics for admitting norms with conflict graphs. The first heuristic, \textit{random drop}, created a conflict-free set by repeatedly selecting a random edge in the conflict graph, then dropping the edge --- this process repeats until the graph is conflict-free. The second heuristic, \textit{maximal conflict-free set}, found the conflict-free set which was maximal wrt.\ the number of norms it contained, with ties being broken randomly. The final heuristic, \textit{preferred extension}, found a preferred extension with the conflict graph, then admitted all norms corresponding to arguments in the extension.

\begin{figure}[!h]
\centering{
    \includegraphics[width=0.7\textwidth]{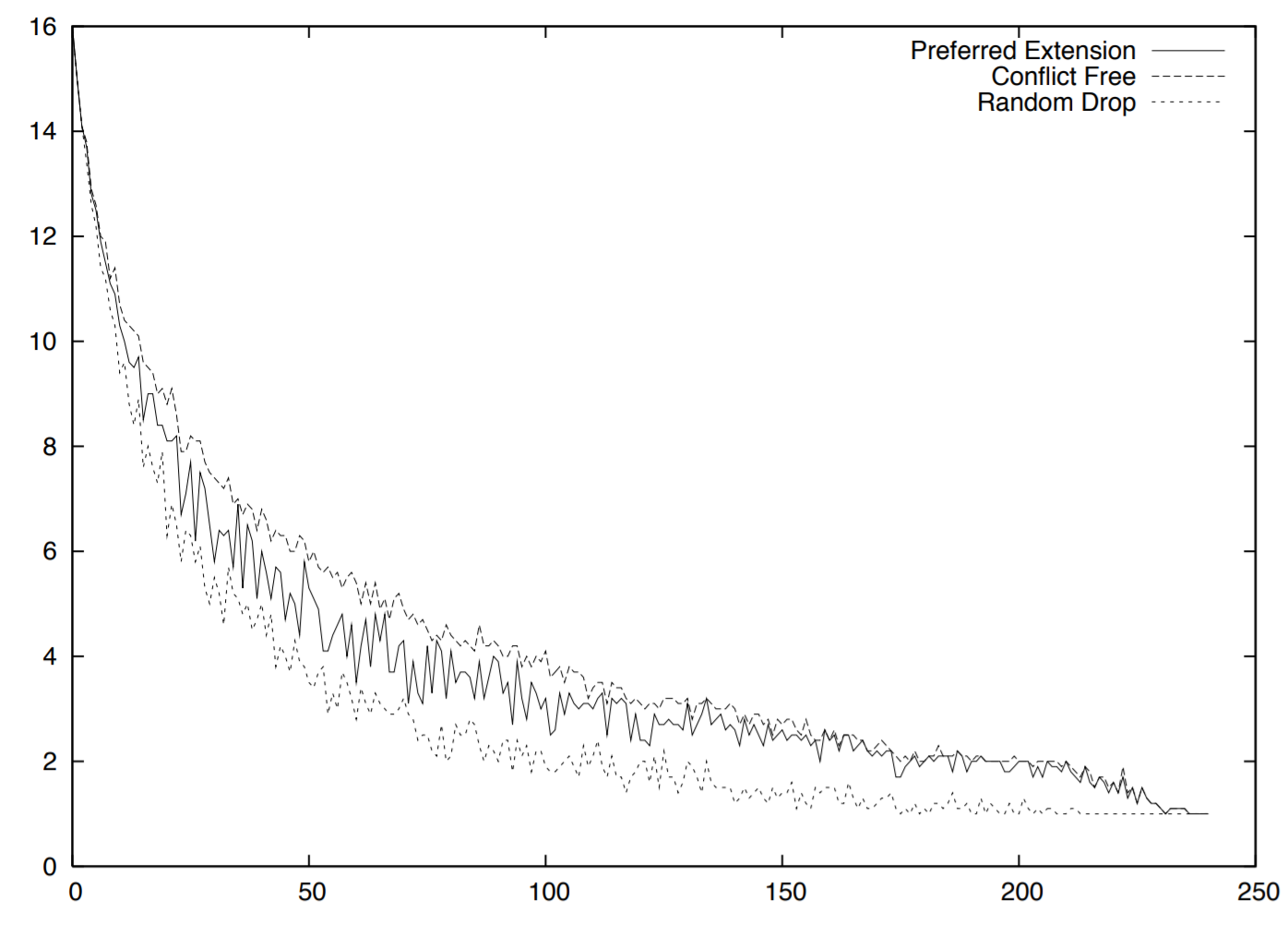}}
    \caption{A direct copy of results of experiments by Oren et al.\ \cite[Fig. 6]{oren2008argumentation}. Description: \textit{``An evaluation of the preferred extension based norm conflict resolution heuristic (solid line), compared to the size of the largest conflict free set (dashed line) and the size of the set of norms obtained when dropping the dominant conflicting norm in random order (dotted line). These results are averaged over 10 runs. The y-axis shows the number of norms retained, while the x-axis indicates the number of normative conflicts." \cite{oren2008argumentation}}}\label{orenfigure}
\end{figure}

These heuristics were evaluated by generating a system of 16 norms. The total number of (directional) conflicts between these norms ranged from 1 to 240 (\footnote{Though not explicitly discussed by Oren et al., 240 is the maximum number of edges possible in a directed graph with 16 edges. An undirected graph with $n$ vertices may have up to ${n \choose 2}$ vertices, so an undirected graph with 16 edges has ${16 \choose 2}=120$ edges. We call a graph with the maximal number of edges a \textit{complete graph}. A complete \textit{directed} graph with 16 vertices has twice as many edges, giving 240 total.}). For each possible number of conflicts, 10 trials were run for each of the three heuristics and the average number of norms admitted over these 10 trials was recorded. The results can be seen in Figure \ref{orenfigure}.

Oren et al.\ also point out that as the number of conflicts increases, the number of norms admitted goes down for all three heuristics.

\subsection{Evaluations on our work and comparison to Oren et al.}\label{sectiontest2}

By following the same methodology as Oren et al., we can obtain a direct comparison between the performances of $ColourCurtail$ and $ColourCurtailComplete$ and the performances of the heuristics proposed by Oren et al.

Although Oren et al.\ use directed graphs and our algorithms are built for undirected graphs, we can still obtain a direct comparison by allowing directed edges and treating them as though they were undirected. This way, we can go up to 240 edges in our graph. All other hyperparameters for our trials will remain the same as with Oren et al.; the graph will always have 16 vertices, edges will be randomly picked, and each number of edges will be trialled 10 times for each algorithm before the average is taken.

\begin{figure}[!h]
\centering{
    \includesvg[width=0.7\textwidth]{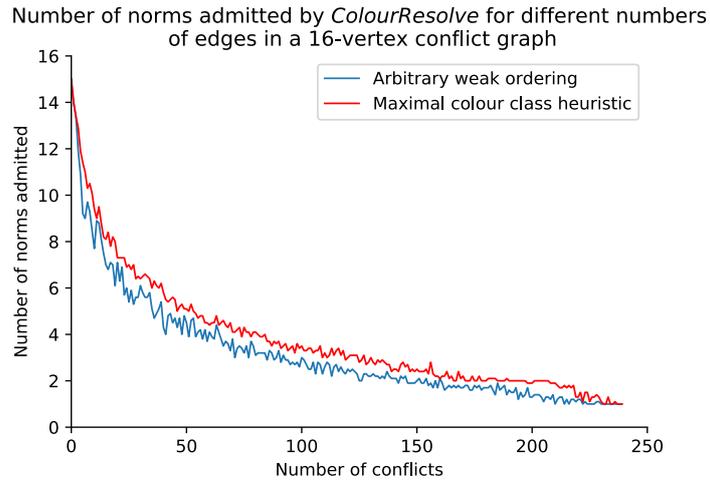}}
    \caption{The number of norms $ColourResolve$ when following the methodology of Oren et al.\ \cite{oren2008argumentation}. The blue line represents the performance under a heuristic which considers an arbitrary weak ordering (which generalises lex superior, lex posterior, and lex specialis). The red line represents the performance when selecting norms corresponding to the largest colour class.}\label{ccresults}
\end{figure}

First, let us consider $ColourCurtail$. In Section \ref{sectionheuristics}, we showed that the policies of lex superior, lex posterior, and lex specialis can all be characterised in our algorithm by an arbitrary weak ordering. As such, we will consider the performance of our algorithm for both a weak ordering (Def. \ref{defgeneral}) and a heuristic which selects the largest colour overall class (Def. \ref{defhnum}). Without loss of generality, a weak ordering was defined over the 16 norms by assigning each norm a unique integer representing their value from 1 to 16. We will use Br\'elaz's DSatur algorithm \cite{brelaz1979new} (Algorithm \ref{dsaturalg}) for graph colouring. By following this methodology and the methodology of Oren et al.\ we obtain results which can be seen in Figure \ref{ccresults}

\begin{figure}[!h]
\centering{
    \includegraphics[width=0.7\textwidth]{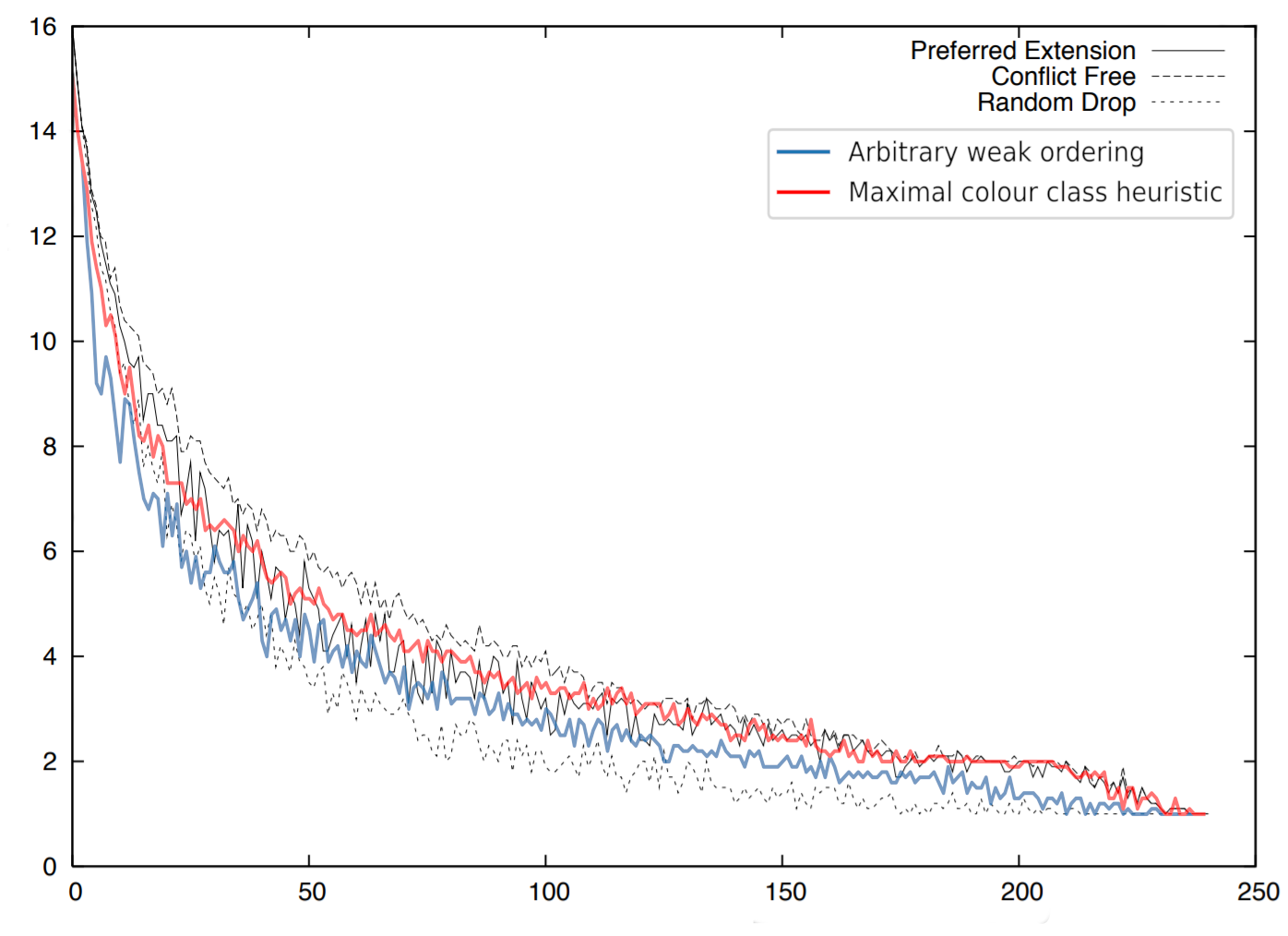}}
    \caption{The results of the evaluations Oren et al.\ \cite{oren2008argumentation} with the results of our evaluations (Figure \ref{ccresults}) overlaid directly.}\label{overlay1}
\end{figure}

As with Figure \ref{orenfigure}, we can see in Figure \ref{ccresults} that as the number of conflicts increases, the number of norms admitted decreases as expected. To obtain a direct comparison between our results and those of Oren et al., we can overlay Figures \ref{orenfigure} and \ref{ccresults} directly --- the results of doing so can be seen in Figure \ref{overlay1} \footnote{This was achieved by plotting our results over the same intervals as Oren et al.\ in each axis. GIMP, a photo manipulation program, was then used to scale raster images of the axes directly on top of one another before duplicate elements were erased.}.

Figure \ref{overlay1} shows us that $ColourResolve$ outperforms the random drop heuristic in terms of number of norms admitted both when using a weak ordering and the maximal colour class heuristic. The amount of norms admitted when using the maximal colour class heuristic appears to be approximately on par with the maximal conflict-free set heuristic by Oren et al. This is encouraging as it shows that our use of graph colouring has not restricted us to only being able to select sets which are significantly smaller than the largest independent set (though as discussed in Section \ref{sectioncompleteextension}, there are cases where restrictions exist). When using weak orderings, $ColourResolve$ was marginally outclassed by the maximal conflict-free set heuristic in terms of number of norms admitted --- however, this is to be expected since the weak ordering heuristic's strength comes from being able to admit the \textit{most important} norms rather than the greatest overall number of norms. We can also see that the weak ordering and maximal colour class heuristic were both outperformed by the preferred extension heuristic, which as explained earlier in this section, is wholly expected

\begin{figure}[!h]
\centering{
    \subfigure[]{\includesvg[width=0.52\textwidth]{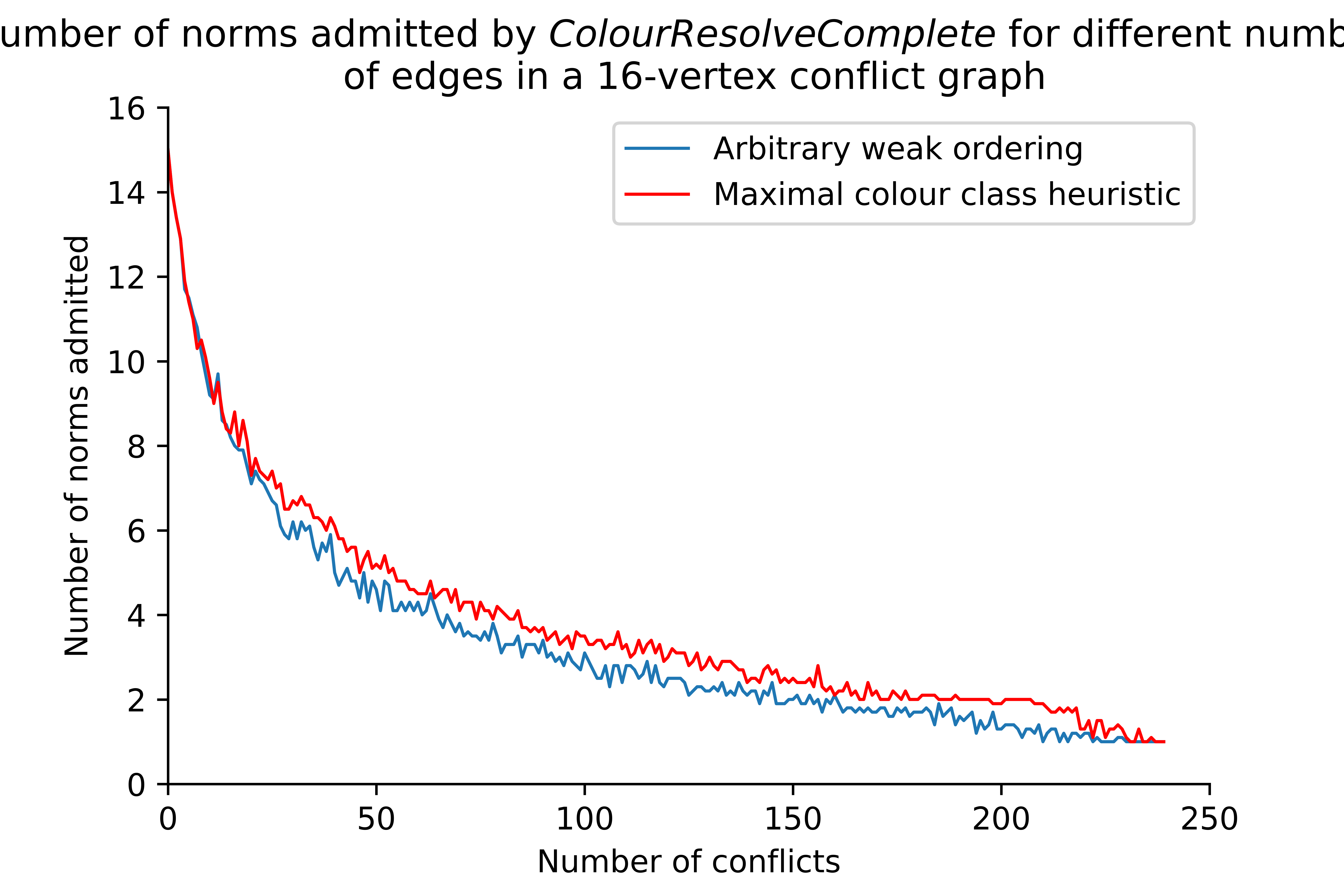}}}\; \; \;
    \subfigure[]{\includegraphics[width=0.4\textwidth]{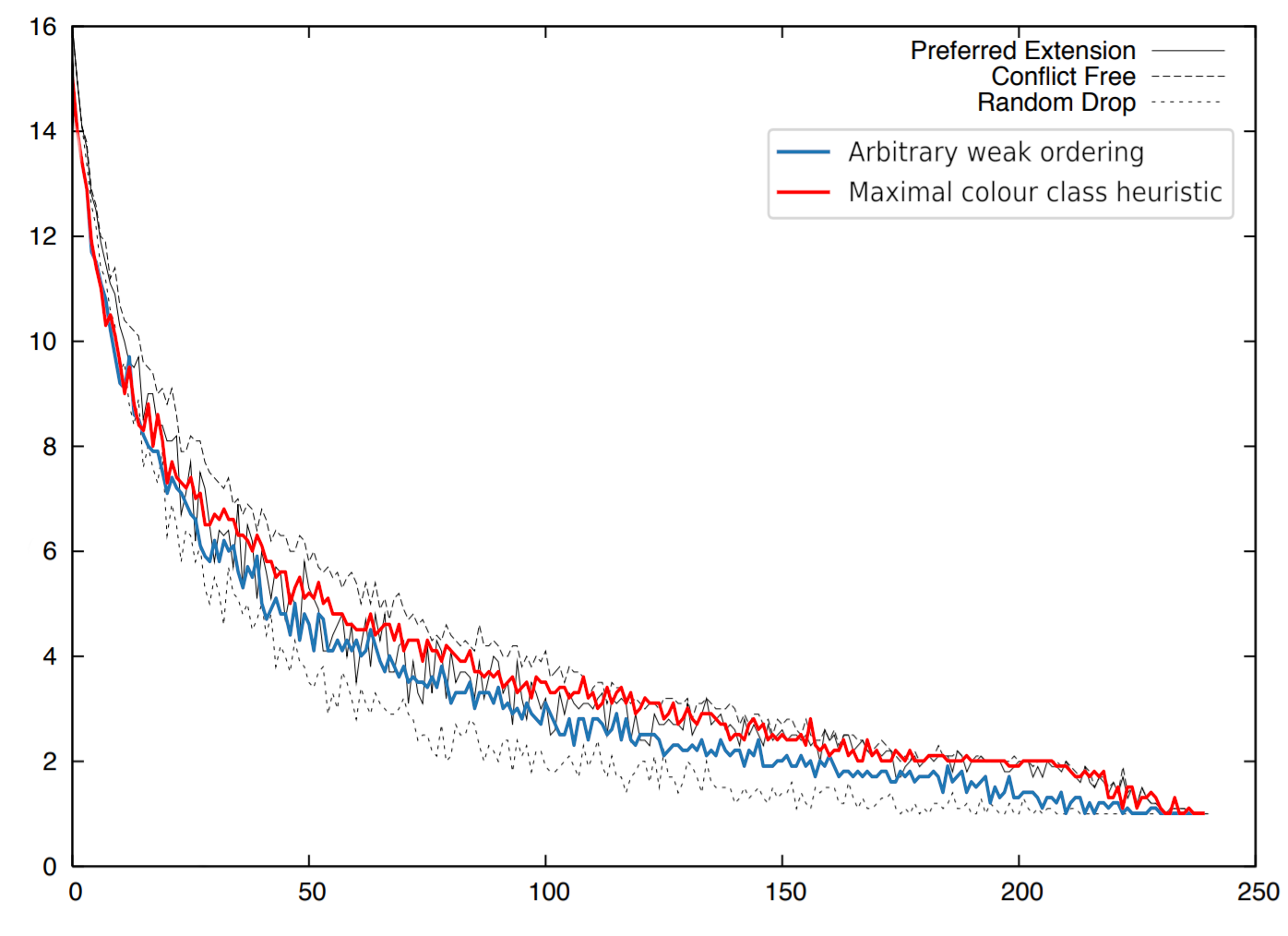}}

    \caption{(a) The number of norms admitted by $ColourCurtailComplete$ when following the methodology of Oren et al. (b) The results of the evaluations Oren et al.\ \cite{oren2008argumentation} with the graph from (a) overlaid directly.}\label{overlay2}
\end{figure}

Let us now consider the performance of $ColourResolveComplete$. We know that $Colour\allowbreak Resolve\allowbreak Complete$ will always admit either the same amount of norms or more than $Colour\allowbreak Resolve$ when given the same heuristic. By following the same evaluation procedures, we obtain results which can be seen in Figure \ref{overlay2}. In particular, Figure \ref{overlay2}(b) shows these results laid over those of Oren et al., as we did earlier. Again, we see that $ColourCurtail$ performs roughly similarly to the maximal conflict-free set heuristic when we use the maximal colour class heuristic. However, a key difference compared to $ColourResolve$ is that when we use a weak ordering as a heuristic, the number of norms we admit is roughly similar to the amount admitted under the maximal conflict-free set heuristic. Therefore, we can see that by ``completing" our set of admitted norms, the number of newly added norms is greatest when we use a weak ordering, so heuristics based on weak orderings benefit the most from completion.

\subsection{Further evaluation}\label{sectiontest3}

\begin{figure}[!h]
\centering{
    \includesvg[width=0.7\textwidth]{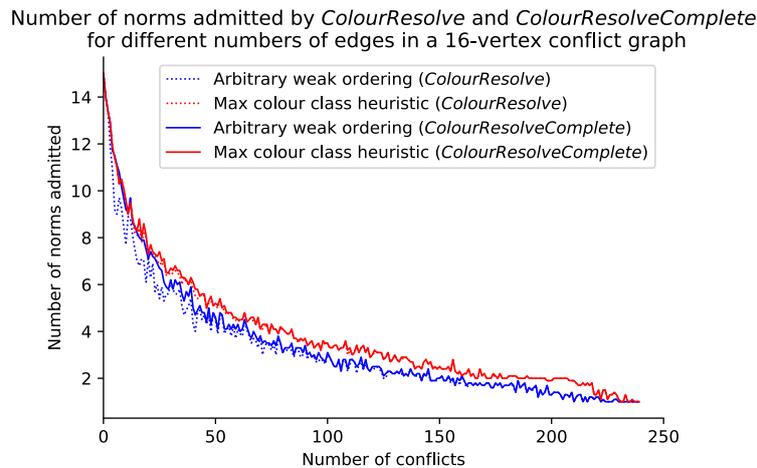}}
    \caption{Direct comparison of Figures \ref{ccresults} (performance of $ColourCurtail$) and \ref{overlay2}(a) (performance of $ColourCurtailComplete$). Colours from these figures have been used, but lines from Figure \ref{ccresults} have been changed to dashed lines.}\label{results-comparison}
\end{figure}

We have so far compared our results to those of Oren et al.\ by using the same methodology. We can also do some brief comparisons between the algorithms we have presented so far. Figure \ref{results-comparison} shows a direct comparison between the performances of $ColourCurtail$ and $Colour\allowbreak Curtail\allowbreak Complete$. The number of norms admitted is roughly the same for each algorthim under the maximal colour class heuristic, which serves to reaffirm our hypothesis that completing the admitted set is most beneficial when using a weak ordering. We can also see that the benefits of doing so are most pronounced when we have fewer conflicts between norms --- this is to be expected, since it is easier to append more norms to the admitted set if there are fewer conflicts preventing us from doing so.

We can also observe that the maximal colour class heuristic appears to admit approximately the same number of norms under $ColourResolve$ and $ColourResolveComplete$, meaning that fewer extra norms are admitted by completing the colour class. This implies that, in practice, the largest colour class in a graph is \textit{often} equal to the maximal independent set of vertices despite this not being a mathematical certainty.

Our evaluation has so far focused on the number of norms admitted by each algorithm. However, this metric neglects the \textit{importance} of each norm when we use a weak ordering. We can evaluate this by using the same methodology we have been using throughout this section, though instead of counting the number of vertices admitted, we will count the number of times each norm we admit is \textit{more preferred} than a conflicting norm which was not admitted. In other words, we will measure the value of the arbitrary weak ordering heuristic over the set of norms we admit. Further to this, the number of times each vertex in the set is \textit{less preferred} than a conflicting vertex, we will subtract 1 from the heuristic's value. This way, the sum of the heuristic values over all norms is 0, so we can judge the worth of an admitted set of norms by how much greater is value is than 0.  

\begin{figure}[!h]
\centering
    \subfigure[]{\includesvg[width=0.49\textwidth]{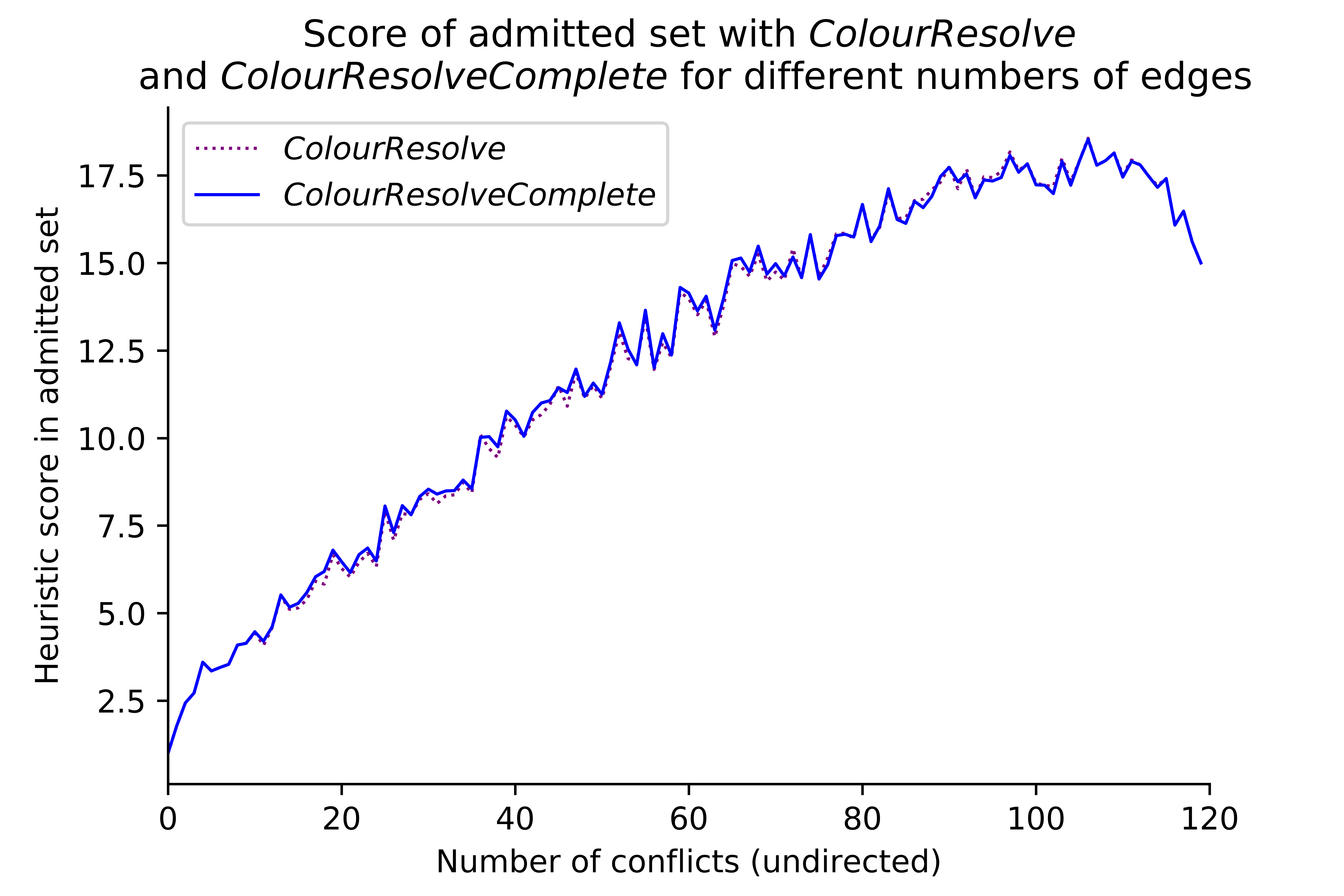}}
    \subfigure[]{\includesvg[width=0.49\textwidth]{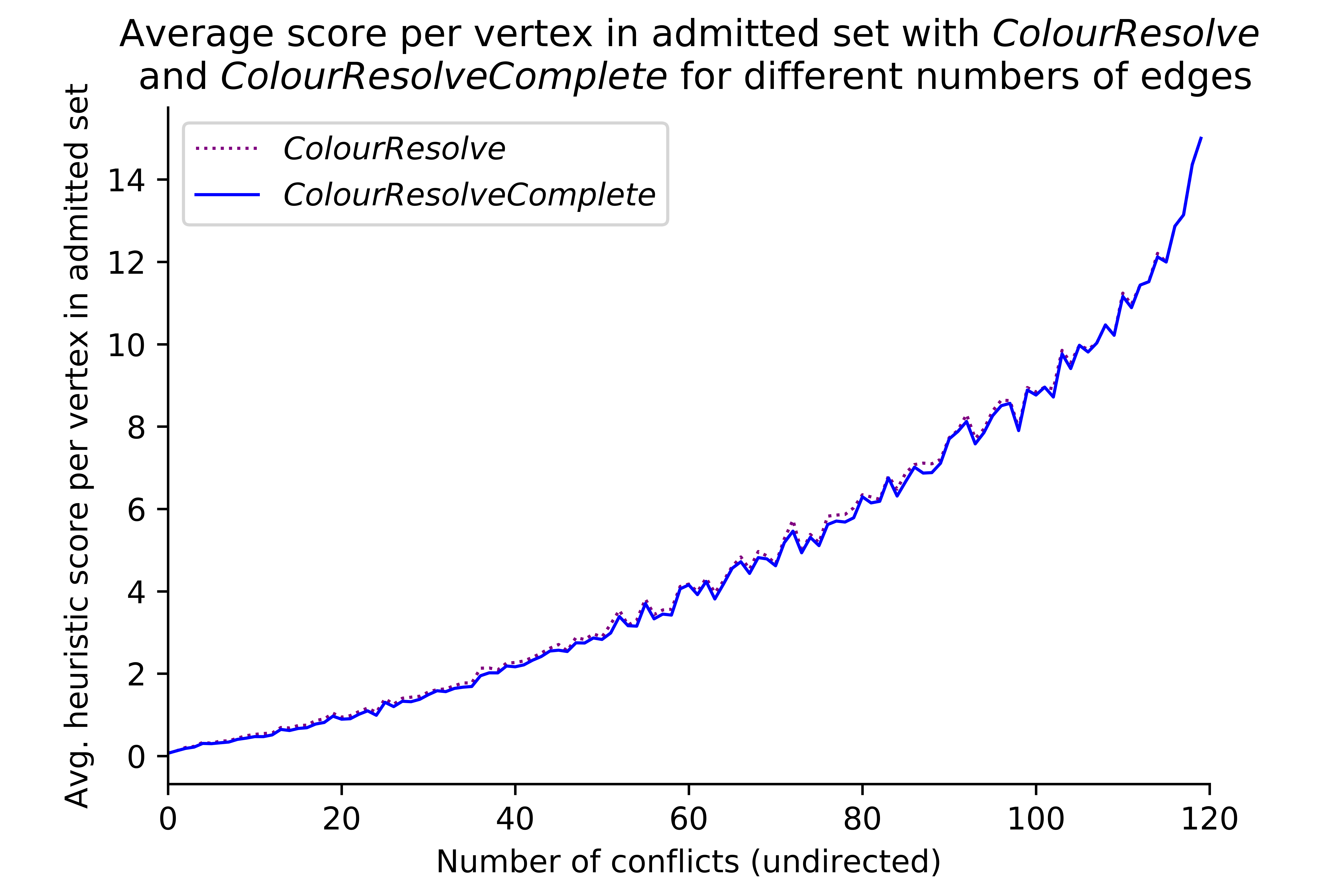}}
    \caption{Plots of the heuristic scores of the admitted sets created by $Colour\allowbreak Resolve$ and $Colour\allowbreak Resolve\allowbreak Complete$ (both under a weak ordering) for different numbers of undirected edges in a conflict graph with 16 vertices. (a) shows the sum of scores of admitted vertices while (b) shows the average score per admitted vertex. Duplicate edges were not allowed, and 250 trials were run per algorithm per number of edges.}\label{results-weakorder}
\end{figure}

Since we are no longer following the methodology of Oren et al., we will no longer allow two edges between each vertex in the conflict graph and we will run 250 trials per number of edges rather than 10 trials.  The results of measuring the heuristic values for $ColourResolve$ and $ColourResolveComplete$ can be seen in Figure \ref{results-weakorder}(a). Additionally, Figure \ref{results-weakorder}(b) shows the same results, except where the score of the admitted set is measured as an average over the admitted vertices.

Surprisingly, we can see that using $ColourResolveComplete$ rather than $ColourResolve$ did not result in a change in either the per-vertex score or the overall score of the set of admitted norms. We can also see from Figure \ref{results-weakorder}(b) that the average score per vertex approaches 15 as the number of conflicts approaches its maximum of 120. This is as expected since when we have 120 conflicts, every norm conflicts with every other norm. The heuristic will therefore choose the most preferred norm in the whole set --- this outranks every other norm in the weak ordering and therefore scores 15 by scoring 1 in each of its conflicts. On the other hand, when there are very few conflicts, the average score per vertex is near 0. This is to be expected since in this case we are likely to admit a large number of norms, only a few of which are involved in any conflicts.

\subsubsection{A note on empirical evaluation of $ColourCurtail$ and $ColourCurtailComplete$}

Although $ColourCurtail$ and $ColourCurtailComplete$ were not evaluated, their performances outclass those of $ColourResolve$ and $ColourResolveComplete$ respectively (see Section \ref{sectionnotecomparison}). However, the introduction of curtailment means that we cannot use similar methodology to Oren et al.\ since the question of a norm's admittance is no longer binary. Rather, we choose the extent to which a norm should be curtailed. This requires more complex and nuanced evaluation, which we will not include in this paper. However, evaluating the performances of $ColourCurtail$ and $ColourCurtailComplete$ relative to other methods of resolving normative conflicts may be a potential avenue for future research.

\section{Further work}\label{sectionfurtherwork}

The framework we have introduced opens up various avenues for further work. For example, one could attempt to create a similar system where instead of following a heuristic equivalent a policy for normative conflict resolution, the system would resolve conflicts by using semantics and criteria within the field of argumentation. Such criteria may include the \textit{last link} or \textit{weakest link} criteria for defeating arguments \cite{liao2016prioritized}, as discussed in Section \ref{sectionrelatedarg}. It may also be interesting to attempt to find other heuristics which cannot be characterised as a weak ordering (as with the maximal colour class heuristic). Trust-based heuristics \cite{parsons2011using,keung2008using} may fall under this category.

Alternatively, one could consider a system similar to ours which relies on a conflict graph, though where the edges in the conflict graphs are \textit{directed} --- that is, one argument could attack or conflict with another without the other attacking back. This would allow for graph pruning algorithms, such as those put forth by Oren et al.\ \cite{oren2008argumentation}, to be used. These would reduce the number of normative conflicts and therefore reduce the number of edges in the conflict graph. Having fewer edges in the graph generally allows us to use fewer colours when colouring the graph and therefore allows more arguments to be admitted.

One could also investigate whether the algorithms presented in this paper could be modified to produce a preferred extension or a stable extension (detailed in \cite{dung1995acceptability}). It may be possible to produce these types of extensions by using a specific graph colouring algorithm which prioritises a specific colour class and therefore assigns as many vertices as possible to this colour --- one could then use the $h_{\max}$ heuristic to select this large colour class. The challenge here arises from guaranteeing that this large colour class is a \textit{maximal} independent set. For a further challenge, one could try to create a colouring which corresponds to a \textit{stable extension}, where each argument which is not in the extension is attacked by an argument which is in the extension. To achieve this, one would have to ensure that each vertex either belongs to a given colour class or is connected to a vertex with the given colour class via an edge whilst remaining a valid colouring.

Another area for investigation is how the order in which vertices are considered in \textit{ColourResolveComplete} (Algorithm \ref{variantcompletealg}) affects \textit{which} complete extension is found. For example, one could raise the question of whether a specific heuristic generate a preference ordering which always results in a preferred or stable extension.

For many of the challenges listed in this section, it may be prudent to investigate the link between \textit{maximum independent sets} in graphs and graph colouring. For this, the work of Eppstein \cite{eppstein2002small}, which involves finding exact graph colourings by listing independent sets, may be useful. It may also be useful to investigate preference-based graph colouring, as with Koseki et al.\ \cite{koseki2002preference}. Other avenues may include using more robust graph colouring algorithms, where methods such as those discussed by Lewis et al.\ \cite{lewis2012wide} could be used.

\section{Conclusion}\label{sectionconclusion}

We have used normative reasoning as the baseline for our problem with the goal of resolving normative conflicts. From there, we applied argumentation in the style of Dung \cite{dung1995acceptability} and conflict graphs in the style of Oren et al.\ \cite{oren2008argumentation}, allowing us to use results from the field of graph theory. In our case, we applied graph colouring and briefly showed the efficient and easily comprehensible \textit{DSatur} graph colouring algorithm by Br\'elaz \cite{brelaz1979new}.

We introduced the notion of a heuristic function to assign a value to each conflict-free set of norms corresponding to each colour class, allowing us to introduce our first algorithm, \textit{ColourResolve}, in Algorithm \ref{mainalg}. We discussed the versatility and shortcomings of \textit{ColourResolve} and showed in Proposition \ref{propadmissible} that \textit{ColourResolve} always produces an admissible set. We showed some possible heuristics one could use with \textit{ColourResolve} and demonstrated the equivalence of heuristics based off of weak orderings when using our algorithm. We then introduced \textit{ColourResolveComplete} (Algorithm \ref{variantcompletealg}), which we proved would always produce a complete extension.

We then discussed the notion of curtailment in the style of Vasconcelos et al.\ \cite{vasconcelos2009normative}, allowing us to create a preference ordering over colours and admit norms corresponding to more than one colour in \textit{ColourCurtail} (Algorithm \ref{curtailalg}). By combining this with the logic of \textit{ColourResolve}, we introduced our most robust algorithm, \textit{ColourCurtailComplete} (Algorithm \ref{curtailcompletealg}), which has computational complexity $\mathcal{O}(n^3)$. \textit{ColourCurtailComplete} finds a graph colouring, then generates an order of preferences over norms corresponding to each colour class according to some heuristic. It then iterates over each colour class in descending order of preference and admits as many vertices as possible into the given colour class. It then uses curtailment to resolve any conflicts between the current colour class and any colour classes which have been admitted so far, then admits the colour class into the resulting set of norms.

Empirical evaluations were performed to compare $ColourResolve$ and $Colour \allowbreak Resolve \allowbreak Complete$ both to each other and to heuristics by Oren et al.\ \cite{oren2008argumentation}. We observed that, in practice, $Colour\allowbreak Resolve$ and $Colour\allowbreak Resolve\allowbreak Complete$ performed approximately similar to choosing a maximal conflict-free set despite their theoretical limitations. We also observed that when we use a weak ordering as a heuristic, $ColourResolve$ and $ColourResolveComplete$ achieve approximately the same heuristic score for admitted vertices. 

In terms of our \textbf{original goals} we outlined in Section \ref{sectionintroduction} and refined in Section \ref{sectionlitreview}, we have successfully created a system which is \textbf{compatible with the policies of lex superior, lex posterior, and lex specialis}, and we have successfully ensured that the results \textbf{align with argumentation-based notions} (in our case, complete extensions). By using the notion of curtailment, we were also able to successfully \textbf{admit norms which directly conflict with one another}, allowing all norms to be admitted in some form.

The results put forth in this paper could potentially be used in their own right or could be used to create more complex methods of conflict resolution via graph colouring. In terms of direct applications, single-agent or multi-agent systems which use norms may benefit from the algorithms proposed in this paper, including models for safe reinforcement learning, real-world intelligent agents, or systems aiming to uphold safety constraints. There may also be further avenues for research into the links between argumentation, normative reasoning, and graph theory. Other graph-theoretic notions involving independent sets may hold interesting results since they necessarily translate into conflict-free sets of norms.

\newpage

\bibliography{bibliography.bib}

\newpage

\appendix

\section{Copy of code used}

The Python code shown below was used to run trials with $ColourResolveComplete$. Equivalent trials were run with $ColourResolve$ by removing the block of code which ``completes" the admitted set. 
% I (John Joyce) wrote all of the code which was used in testing the new algorithms put forth in this paper.

\begin{lstlisting}
import numpy as np

norms = [0,1,2,3,4,5,6,7,8,9,10,11,12,13,14,15]
values = [15,14,13,12,11,10,9,8,7,6,5,4,3,2,1,0]
n = len(norms)
np.random.seed(178231423)

numconflicts = 30
conflicts = []
while len(conflicts) < numconflicts:
    conflict = [np.random.choice(norms),np.random.choice(norms)]
    if conflict[0] != conflict[1]:
        if conflict not in conflicts:
            conflicts.append(conflict)
    
print(conflicts)

def lexposterior(norms,whendeclared,conflict):
    # Evaluate a norm according to arbitrary weak orderings
    index0 = norms.index(conflict[0])
    index1 = norms.index(conflict[1])
    if whendeclared[index0] > whendeclared[index1]:
        return 0
    elif whendeclared[index1] > whendeclared[index0]:
        return 1
    elif index0 > index1:
        return 0
    else:
        return 1
    
def colour_dsatur(vertices,edges):
    # colour a graph using dsatur algorithm
    
    colours = [None for i in range(len(vertices))]
    highestcolour = 0
    n = len(vertices) # number of vertices
    
    # get degrees of each vertex
    degrees = []
    flatedges = [i for x in edges for i in x]
    for i in range(len(vertices)):
        degree = 0
        for j in flatedges:
            if i==j:
                degree += 1
        degrees.append(degree)
    
    # find max-degree vertex
    maxdegree = 0
    maxdegreevertex = None
    for i in range(len(degrees)):
        if maxdegree < degrees[i]:
            maxdegree = degrees[i]
            maxdegreevertex = vertices[i]
    colours[maxdegreevertex] = 0
    
    while None in colours: # while not all vertices have been coloured
        
        # find a vertex with maximum degree of saturation
        dsaturlist = [None for i in range(n)]
        for i in range(n): # consider each vertex
            if colours[i] == None: # if current vertex is not yet coloured
                
                connectedcolours = []
                for edge in edges: # consider each edge
                    
                    if vertices[i] in edge: # if current vertex is in the edge
                        if edge[0] == vertices[i]: # consider colour of the other vertex in the edge
                            othercolour = colours[edge[1]]
                        else:
                            othercolour = colours[edge[0]]
                        if othercolour != None: # consider colour of other vertex
                            if othercolour not in connectedcolours:
                                connectedcolours.append(othercolour) # count colour of other vertex
                dsaturlist[i] = len(connectedcolours) # save degree of saturation of current vertex
        
        maxdsatur = max([i for i in dsaturlist if i != None]) # choose a vertex with maximal dsatur. break ties by choosing highest degree vertex
        maxdegree = 0
        for i in range(n):
            if dsaturlist[i] == maxdsatur and degrees[i] > maxdegree and colours[i] == None:
                maxdegree = degrees[i]
                vertextocolour = i
        if maxdegree == 0: # if the only remaining vertices are isolated
            for i in range(n): # just pick the isolated vertices in order of index
                if colours[i] == None:
                    vertextocolour = i
                    break
        
        # assign the vertex the lowest possible colour
        neighbouringcolours = []
        for edge in edges:
            if vertices[vertextocolour] in edge:
                if vertices[vertextocolour] == edge[0]:
                    othervertex = edge[1] # get other vertex in edge
                else:
                    othervertex = edge[0]
                if colours[othervertex] != None: # if neighbouring vertex has been coloured
                    if colours[othervertex] not in neighbouringcolours: # append colour to list of neighbouring colours
                        neighbouringcolours.append(colours[othervertex])
        for c in colours: # loop through each colour in order
            if c != None:
                if c not in neighbouringcolours: # check if colour can be used
                    colours[vertextocolour] = c
                    break
        else: # if no colour can be used
            colours[vertextocolour] = max([i for i in dsaturlist if i != None]) # create new colour
    return colours


def getcolourscores(vertices,edges,colouring):
    colours = list(range(max(colouring)+1))
    colourvalues = [0 for i in colours]

    for c in colours: # go through each colour
        for edge in conflicts: # find winner for each edge
            if colouring[edge[0]] == c or colouring[edge[1]] == c:
                winner = lexposterior(norms,values,edge)
                if winner == 0 and colouring[edge[0]] == c:
                    colourvalues[colours.index(c)] += 1
                elif winner == 1 and colouring[edge[0]] == c:
                    colourvalues[colours.index(c)] -= 1
                elif winner == 1 and colouring[edge[1]] == c:
                    colourvalues[colours.index(c)] += 1
                else:
                    colourvalues[colours.index(c)] -= 1
    return colourvalues
    
    
'''
RUN TRIALS USING ARBITRARY WEAK ORDERING (i.e. lex superior, lex posterior, lex specialis, etc.)
'''


norms = [0,1,2,3,4,5,6,7,8,9,10,11,12,13,14,15]
values = [15,14,13,12,11,10,9,8,7,6,5,4,3,2,1,0]
n = len(norms)
np.random.seed(178231423)


results = []
for numconflicts in range(1,241): # Perform a run of 10 trials for # of conflicts ranging from 1 to 240
    print("NUMCONFLICTS:", numconflicts)
    
    trialresults = []
    for trial in range(10):
        
         # Generate conflicts
        conflicts = []
        while len(conflicts) < numconflicts:
            conflict = [np.random.choice(norms),np.random.choice(norms)]
            if conflict[0] != conflict[1]:
                if conflict not in conflicts:
                    conflicts.append(conflict)

        # Colour conflict graph
        colouring = colour_dsatur(norms,conflicts)
        colours = list(range(max(colouring)+1))
        
        # Score each colour
        scores = getcolourscores(norms,conflicts,colouring)
        maxcolour = scores.index(max(scores))

        # Get a set of admitted norms
        admitset = []
        for i in range(n):
            if colouring[i] == maxcolour:
                admitset.append(norms[i])
                
        for i in range(n): # COMPLETE the extension. Remove this block of code to turn into ColourResolve
            for edge in conflicts:
                if (norms[i] in edge) and (edge[0] in admitset or edge[1] in admitset):
                    break
            else:
                if norms[i] not in admitset:
                    admitset.append(norms[i])
            
                
        # Save the size of the admitted set as the result of the trial
        trialresults.append(len(admitset))
    
    # Take the average for the entire trial
    results.append(np.mean(trialresults))
    np.save("results-lexposterior-complete.npy",results)
    
print(results)


'''
RUN TRIALS USING MAXIMAL COLOUR CLASS HEURISTIC
'''



norms = [0,1,2,3,4,5,6,7,8,9,10,11,12,13,14,15]
values = [15,14,13,12,11,10,9,8,7,6,5,4,3,2,1,0]
n = len(norms)
np.random.seed(178231423)


results_maxheur = []
for numconflicts in range(1,241): # Perform a run of 10 trials for # of conflicts ranging from 1 to 240
    print("NUMCONFLICTS:", numconflicts)
    
    trialresults = []
    for trial in range(10):
        
         # Generate conflicts
        conflicts = []
        while len(conflicts) < numconflicts:
            conflict = [np.random.choice(norms),np.random.choice(norms)]
            if conflict[0] != conflict[1]:
                if conflict not in conflicts:
                    conflicts.append(conflict)

        # Colour conflict graph
        colouring = colour_dsatur(norms,conflicts)
        colours = list(range(max(colouring)+1))
        
        # Score each colour USING MAXIMAL COLOUR CLASS HEURISTIC
        maxcolour = max(set(colouring), key=colouring.count)

        # Get a set of admitted norms
        admitset = []
        for i in range(n):
            if colouring[i] == maxcolour:
                admitset.append(norms[i])
                
        for i in range(n): # COMPLETE the extension. Remove this code to turn into ColourComplete
            for edge in conflicts:
                if (norms[i] in edge) and (edge[0] in admitset or edge[1] in admitset):
                    break
            else:
                if norms[i] not in admitset:
                    admitset.append(norms[i])
                
        # Save the size of the admitted set as the result of the trial
        trialresults.append(len(admitset))
    
    # Take the average for the entire trial
    results_maxheur.append(np.mean(trialresults))
    
print(results_maxheur)
np.save("results-maxheur-complete.npy",results_maxheur)
\end{lstlisting}

\newpage

After the above code has was run, the code below was used to run trials where the heuristic scores of vertices in the admitted set were measured.

\begin{lstlisting}

'''
RUN TRIALS USING ARBITRARY WEAK ORDERING (i.e. lex superior, lex posterior, lex specialis, etc.)
'''


norms = [0,1,2,3,4,5,6,7,8,9,10,11,12,13,14,15]
values = [15,14,13,12,11,10,9,8,7,6,5,4,3,2,1,0]
n = len(norms)
np.random.seed(178231423)


results = []
for numconflicts in range(1,241): # Perform a run of 10 trials for # of conflicts ranging from 1 to 240
    print("NUMCONFLICTS:", numconflicts)
    
    trialresults = []
    for trial in range(10):
        
         # Generate conflicts
        conflicts = []
        while len(conflicts) < numconflicts:
            conflict = [np.random.choice(norms),np.random.choice(norms)]
            if conflict[0] != conflict[1]:
                if conflict not in conflicts:
                    conflicts.append(conflict)

        # Colour conflict graph
        colouring = colour_dsatur(norms,conflicts)
        colours = list(range(max(colouring)+1))
        
        # Score each colour
        scores = getcolourscores(norms,conflicts,colouring)
        maxcolour = scores.index(max(scores))

        # Get a set of admitted norms
        admitset = []
        for i in range(n):
            if colouring[i] == maxcolour:
                admitset.append(norms[i])
                
        for i in range(n): # COMPLETE the extension. Remove this block of code to turn into ColourResolve
            for edge in conflicts:
                if (norms[i] in edge) and (edge[0] in admitset or edge[1] in admitset):
                    break
            else:
                if norms[i] not in admitset:
                    admitset.append(norms[i])
            
                
        # Save the size of the admitted set as the result of the trial
        trialresults.append(len(admitset))
    
    # Take the average for the entire trial
    results.append(np.mean(trialresults))
    np.save("results-lexposterior-complete.npy",results)
    
print(results)


\end{lstlisting}

% \newpage

% \section{Contents of design archive}

% Contents of the design archive are as follows:

% \begin{itemize}\itemsep0em
%     \item One folder called \textit{``Image files for document"}. This contains various figures used in this paper.
    
%     \item One folder called \textit{``Code and Figures"}. This contains:
    
%     \begin{itemize}\itemsep0em
%         \item A file \textit{``ColourResolve.ipynb"}. This was made in and can be run with Jupyter Notebook. This includes the code which was used to evaluate $ColourResolve$.
%         \item \textit{``ColourResolveComplete.ipynb"}. This was used to evaluate \textit{ColourResolveComplete}.
%         \item \textit{``ColourResolve weak ordering comparison.ipynb"}. This was used to evaluate $ColourResolve$ and $ColourResolveComplete$ via their heuristic scores.
%         \item \textit{``Oren comparison.xcf"}. This contains the layers used to overlay results from Oren et al.\ with our own and can be opened in GIMP, Photoshop, or a similar program. An invisible layer is included showing the two axes overlaid before the redundant axis was erased. Similarly for \textit{``Oren comparison 2.xcf}.
%         \item Miscellaneous .png and .svg files. These were used in images in this paper.
%         \item Miscellaneous .npy files. These were created as outputs by the .ipynb files and they match the outputs shown in appendix B.
%         \item A folder called \textit{``.ipynb\_checkpoints"}. This contains automatically-generated backups of the .ipynb files.
%     \end{itemize}
% \end{itemize}

% %TC:endignore

\end{document}